\documentclass{article} 
\usepackage{nips14submit_e,times}
\usepackage{url}
\usepackage[dvips]{graphicx}
\usepackage[psamsfonts]{amssymb}
\usepackage{amsmath,amsfonts,amsthm,epsfig,epstopdf,url,array}
\usepackage{grffile}
\usepackage{mathtools}
\usepackage{subfig}
\usepackage[linesnumbered,ruled,vlined]{algorithm2e}
\usepackage{caption}

\newcommand{\vect}[1]{\boldsymbol{#1}}
\newcommand{\argmin}{\arg\!\min}

\newcommand{\specialcell}[2][c]{%
  \begin{tabular}[#1]{@{}c@{}}#2\end{tabular}}

\theoremstyle{plain}
\newtheorem{thm}{Theorem}[section]
\newtheorem{lem}[thm]{Lemma}

\theoremstyle{definition}

\theoremstyle{remark}

\newcommand{\diag}[1]{\mbox{\textbf{diag}}(#1)}
\usepackage{bm}
\newcommand{\bs}{\boldsymbol}
\newcommand{\red}[1]{\textcolor{red}{#1}}
\newcommand{\blue}[1]{\textcolor{blue}{#1}}

\title{Distributed Weighted Parameter Averaging for SVM Training on Big Data}

 \author{
 Ayan Das \\
 Dept. of Computer Science and Engineering \\
 IIT Kharagpur, Kharagpur \\
 W.B. - 721032, India \\
 \texttt{ayand@cse.iitkgp.ernet.in} \\
 \And
 Sourangshu Bhattacharya \\
 Dept. of Computer Science and Engineering \\
 IIT Kharagpur, Kharagpur \\
 W.B. - 721032, India \\
 \texttt{sourangshu@cse.iitkgp.ernet.in} \\
 }

\nipsfinalcopy 

\begin{document}
\setlength{\abovedisplayskip}{0pt}
\setlength{\belowdisplayskip}{0pt}
\setlength{\abovedisplayshortskip}{0pt}
\setlength{\belowdisplayshortskip}{0pt}

\maketitle

\begin{abstract}
Two popular approaches for distributed training of SVMs on big data are parameter averaging and ADMM. Parameter averaging is efficient but suffers from loss of accuracy with increase in number of partitions, while ADMM in the feature space is accurate but suffers from slow convergence. In this paper, we report a hybrid approach called weighted parameter averaging (WPA), which optimizes the regularized hinge loss with respect to weights on parameters. The problem is shown to be same as solving SVM in a projected space. We also demonstrate an $O(\frac{1}{N})$ stability bound on final hypothesis given by WPA, using novel proof techniques. Experimental results on a variety of toy and real world datasets show that our approach is significantly more accurate than parameter averaging for high number of partitions. It is also seen the proposed method enjoys much faster convergence compared to ADMM in features space.
\end{abstract}

\section{Introduction}
\label{sec:intro}

With the growing popularity of Big Data platforms \cite{hadoop} for various machine learning and data analytics applications ~\cite{mann09, zinekevich10}, distributed training of Support Vector Machines (SVMs)\cite{vapnik95} on Big Data platforms have become increasingly important. Big data platforms such as Hadoop \cite{hadoop} provide simple programming abstraction (Map Reduce), scalability and fault tolerance at the cost of distributed iterative computation being slow and expensive \cite{mann09}. Thus, there is a need for SVM training algorithms which are efficient both in terms of the number of iterations and volume of data communicated per iteration.

The problem of distributed training of support vector machines (SVM) \cite{forero10} in particular, and distributed regularized loss minimization (RLM) in general \cite{boyd11, mann09}, has received a lot of attention in the recent times. Here, the training data is partitioned into $M$-nodes, each having $L$ datapoints. Parameter averaging (PA), also called ``mixture weights'' \cite{mann09} or ``parallelized SGD'' \cite{zinekevich10}, suggests solving an appropriate RLM problem on data in each node, and use average of the resultant parameters. Hence, a single distributed iteration is needed. However, as shown in this paper, the accuracy of this approach reduces with increase in number of partitions. Another interesting result described in \cite{mann09} is a bound of $O(\frac{1}{ML})$ on the stability of the final hypothesis, which results in a bound on deviation from optimizer of generalization error.

Another popular approach for distributed RLM is \textit{alternating direction method of multipliers} (ADMM) ~\cite{boyd11,forero10}. This approach tries to achieve consensus between parameters at different nodes while optimizing the objective function. It achieves optimal performance irrespective of the number of partitions. However, this approach needs  many distributed iterations. Also, 
number of parameters to be communicated among machines per iteration is same as the dimension of the problem. This can be $\sim$ millions for some practical datasets, e.g. webspam \cite{libsvmdata}.

In this paper, we propose a hybrid approach which uses weighted parameter averaging and proposes to learn the weights in a distributed manner from the data. 
We propose a novel SVM-like formulation for learning the weights of the weighted parameter averaging (WPA) model. The dual of WPA turns out to be same as SVM dual, with data projected in a lower dimensional space. We propose an ADMM \cite{boyd11} based distributed algorithm (DWPA), and an accelerated version (DWPAacc), for learning the weights.

Another contribution is a $O(\frac{1}{ML})$ bound on the stability of final hypothesis leading to a bound on deviation from optimizer of generalization error. This requires a novel proof technique as both the original parameters and the weights are solutions to optimization problems (section \ref{sec:stability}). Empirically, we show that that accuracy of parameter averaging degrades with increase in the number of partitions. Experimental results on real world datasets show that DWPA and DWPAacc achieve better accuracies than PA as the number of partitions increase, while requiring lower number of iterations and time per iteration compared to ADMM. 


\section{Distributed Weighted Parameter Averaging (DWPA)}
\label{sec:DWPA}
\vspace{-2mm}
In this section, we describe the distributed SVM training problem, the proposed solution approach and a distributed algorithm. We describe a bound on stability of the final hypothesis in section \ref{sec:stability}. Note that, we focus on the distributed SVM problem for simplicity. The techniques described here are applicable to other distributed regularized risk minimization problems.
\vspace{-2mm}

\subsection{Background}
\label{sec:background}
\vspace{-2mm}
Given a training dataset $S = \{(\mathbf{x}_i,y_i) : i=1,\cdots,ML$, $y_i \in \{-1,+1\},\mathbf{x_i \in \mathbf{R}^d}\}$, the linear SVM problem \cite{vapnik95} is given by: 
\begin{equation}
\label{eq:nondistribsvm}
\min_{\mathbf{w}} \lambda\|\mathbf{w}\|_2^2+\frac{1}{m}\sum_{i=1}^{ML} loss(\mathbf{w};(\mathbf{x}_i,y_i)),
\end{equation}
where, $\lambda$ is the regularization parameter and
the hinge loss is defined as $loss(\mathbf{w};(\mathbf{x}_i,y_i)) = max(0,1-y_i\mathbf{w}^T\mathbf{x}_i)$.
The separating hyperplane is given by the equation $\mathbf{w}^T\mathbf{x} + b = 0$. Here we include the bias $b$ within
$\mathbf{w}$ by making the following transformation, $\mathbf{w} = \left[\mathbf{w}^T,b\right]^T$ and $\mathbf{x}_i = \left[\mathbf{x}_i^T,1\right]^T$.


The above SVM problem can be posed to be solved in a distributed manner, which is interesting when the volume of training data is too large to be effectively stored and processed on a single computer.
Let the dataset be which is partitioned into $M$ partitions ($\mathcal{S}_m,\ m=1,\dots,M$), each having $L$ datapoints. Hence, $\mathcal{S}=\mathcal{S}_1\cup,\dots,\cup\mathcal{S}_M$, where $\mathcal{S}_m=\{(\mathbf{x}_{ml},y_{ml})\},l=1,\dots,L$.
Under this setting, the SVM problem (Eqn \ref{eq:nondistribsvm}), can be stated as:
\begin{align}
 & \min_{\mathbf{w}_m,\mathbf{z}}  \sum_{m=1}^M \sum_{l=1}^{L} loss(\mathbf{w}_{m};(\mathbf{x}_{ml},y_{ml})) + r(\mathbf{z}) \label{eq:distribprob} \\
 & \mbox{s.t.} \mathbf{w}_{m} - \mathbf{z} = 0, m=1,\cdots,M,\ l=1,\dots,L \nonumber
\end{align}
where $loss()$ is as described above and $r(\mathbf{z}) = \lambda\| \mathbf{z} \|^2$.
This problem is solved in \cite{boyd11} using ADMM (see section \ref{sec:distalgo}). 
 
 
Another method for solving distributed RLM problems, called parameter averaging (PA), was proposed by Mann et. al. \cite{mann09}, in the context of {\it conditional maximum entropy} model. Let 
$\hat{\mathbf{w}}_m = \argmin_{\mathbf{w}} \frac{1}{L}\sum_{l=1}^{L} loss(\mathbf{w};\mathbf{x}_{ml},y_{ml}) + \lambda \|\mathbf{w}\|^2\ ,\ m=1,\dots,M$ be the standard SVM solution obtained by training on partition $\mathcal{S}_m$.
Mann et al. \cite{mann09} suggest the approximate final parameter to be the arithmetic mean of the parameters learnt on individual partitions, ($\hat{\mathbf{w}}_m$). Hence:
\begin{align}
 & \mathbf{w}_{PA} = \frac{1}{M}\sum_{m=1}^M \hat{\mathbf{w}}_m \label{eq:PA}
\end{align}
Zinekevich et al. \cite{zinekevich10} have also suggested a similar approach where $\hat{\mathbf{w}}_m$'s are learnt using SGD.
We tried out this approach for SVM. Note that assumptions regarding differentiability of loss function made in \cite{boyd11} can be relaxed in case of convex loss function with an appropriate definition of bregmann divergence using sub-gradients (see \cite{mohribook}, section \ref{sec:stability}). The results (reported in section \ref{sec:exp}) show that the method fails to perform well as the number of partitions increase.
This drawback of the above mentioned approach motivated us to propose the weighted parameter averaging method described in the next section.

\newcommand{\RR}{\mathbb{R}}

\subsection{Weighted parameter averaging (WPA)}
\label{sec:formulation}
\vspace{-2mm}

The parameter averaging method uses uniform weight of $\frac{1}{M}$ for each of the $M$ components. One can conceive a more general setting where the final hypothesis is a weighted sum of the parameters obtained on each partition: $\mathbf{w} = \sum_{m=1}^M\beta_m\mathbf{\hat{w}_m}$, where $\mathbf{\hat{w}_m}$ are as defined above and $\beta_m \in \RR, m=1,\dots,M$. Thus, $ \bm{\beta} = [\beta_1,\cdots,\beta_M]^T = [\frac{1}{M},\dots,\frac{1}{M}]$ achieves the PA setting. Note that Mann et al. \cite{mann09} proposed $\bm{\beta}$ to be in a simplex. However, no scheme was suggested for learning an appropriate $\bm{\beta}$.

Our aim is to find the optimal set of weights $\bm{\beta}$ which attains the lowest regularized loss.
Let $\mathbf{\hat{W}} = [\hat{\mathbf{w}_1},\cdots,\hat{\mathbf{w}_M}]$, so that $\mathbf{w} = \mathbf{\hat{W}}\bm{\beta}$.
Substituting $\mathbf{w}$ in eqn. \ref{eq:nondistribsvm}, the regularized loss minimization problem becomes:
\begin{align}
&  \min_{\bm{\beta}, \mathbf{\xi}} \lambda \|\mathbf{\hat{W}} \bm{\beta}\|^2 + \frac{1}{ML} \sum_{m=1}^{M} \sum_{i=1}^{l} \xi_{mi} \label{eq:nondistopt} \\
& \mbox{\textbf{subject to:} } y_{mi}(\bm{\beta}^T \mathbf{\hat{W}}^T \mathbf{x}_{mi}) \geq 1 - \xi_{mi}, \hspace{0.25cm} \forall i,m \nonumber \\
& \hspace{2cm} \xi_{mi} \geq 0, \hspace{0.25cm} \forall m=1,\dots,M,\ i=1,\dots,l \nonumber
\end{align}
Note that, here the optimization is only w.r.t. $\bm{\beta}$ and $\xi_{m,i}$. $\hat{\bm{W}}$ is a pre-computed parameter. Next we can derive the dual formulation by writing the lagrangian and eliminating the primal variables. The Lagrangian is given by:
\begin{align}
\scriptsize{
 \mathcal{L}(\bm{\beta}, \xi_{mi}, \alpha_{mi}, \mu_{mi}) = \lambda \|\mathbf{\hat{W}} \bm{\beta}\|^2 + \frac{1}{ML} \sum_{m,i}\xi_{mi} + \sum_{m,i} \alpha_{mi} ( y_{mi}(\bm{\beta}^T W^T \mathbf{x}_{mi}) - 1 + \xi_{mi}) - \sum_{m,i} \mu_{mi} \xi_{mi}
 } \nonumber
\end{align}
Differentiating the Lagrangian w.r.t. $\bm{\beta}$ and equating to zero, we get:
\begin{equation}
\label{beta:dual}
\bm{\beta} = \frac{1}{2\lambda}(\mathbf{\hat{W}}^T\mathbf{\hat{W}})^{-1}(\sum_{m,i}\alpha_{mi}y_{mi}\mathbf{\hat{W}}^T\mathbf{x}_{mi})
\end{equation}
Differentiating $\mathcal{L}$ w.r.t. $\xi_{mi}$ and equating to zero, $\forall i \in {1,\cdots, L}$ and $\forall m \in {1,\cdots,M}$, implies $\frac{1}{ML} - \alpha_{mi} - \mu_{mi} = 0 $.
Since $\mu_{mi} \geq 0$ and $\alpha_{mi} \geq 0$, $0 \leq \alpha_{mi} \leq \frac{1}{ML} $. 
Substituting the value of $\bm{\beta}$ in the Lagrangian $\mathcal{L}$, we get the dual problem:
\begin{align}
& \min_{\bm{\alpha}}\mathcal{L}(\bm{\alpha}) = \sum_{m,i}\alpha_{mi} - \frac{1}{4\lambda}\sum_{m,i}\sum_{m',j}\alpha_{mi}\alpha_{m'j}y_{mi}y_{m'j}(\mathbf{x}_{mi}^T\mathbf{\hat{W}}(\mathbf{\hat{W}}^T\mathbf{\hat{W}})^{-1}\mathbf{\hat{W}}^T\mathbf{x}_{m'j}) \label{dual:formula}\\
& \mbox{\textbf{subject to:}  } 0 \leq \alpha_{mi} \leq \frac{1}{ML}\ \ \ \forall i \in {1,\cdots, L}, m \in {1,\cdots,M} \nonumber
\end{align}
Note that this is equivalent to solving SVM using the projected datapoint $(\mathcal{H}\mathbf{x}_{mi},y_{mi})$, instead of $(\mathbf{x}_{mi},y_{mi})$, where $\mathcal{H}=\mathbf{\hat{W}}(\mathbf{\hat{W}}^T\mathbf{\hat{W}})^{-1}\mathbf{\hat{W}}^T$, which is the projection on column space of $\mathbf{\hat{W}}$. Hence the performance of the method is expected to depend on size and orientation of the column space of $\mathbf{\hat{W}}$. Next, we describe distributed algorithms for learning $\bm{\beta}$.

\vspace{-2mm}

\subsection{Distributed algorithms for WPA using ADMM}
\label{sec:distalgo}
\vspace{-2mm}

In the distributed setting, we assume the presence of a central (master) computer which stores and updates the final hypothesis. The partitions of training set $\mathcal{S}_1,\dots,\mathcal{S}_M$ are distributed to $M$ (slave) computers, where the local optimizations are performed. The master needs to communicate to slaves and vice versa. However, no communication between slaves is necessary. Thus, the underlying networks has a star topology, which is also easily implemented using Big data platforms like Hadoop \cite{hadoop}.

Let $\bm{\gamma}_m$, for $m = 1,\cdots,M$ be the weight values at the $M$ different nodes and $\bm{\beta}$ be the value of the weights at the central server. The formulation given in eqn. \ref{eq:nondistopt} can be written as:
\begin{align}
 & \min_{\mathbf{\gamma}_m,\mathbf{\beta}}  \frac{1}{ML} \sum_{m=1}^M \sum_{l=1}^{L} loss(\hat{W}\bm{\gamma}_m;\mathbf{x}_{ml},y_{ml}) + r(\bs{\beta}) \label{eq:distribprobadmm} \\
 & \mbox{s.t.}\hspace{0.5cm} \bs{\gamma}_{m} - \bs{\beta} = 0,\hspace{0.25cm} m=1,\cdots,M,\ \nonumber
\end{align}
where $r(\bs{\beta})=\lambda\|\hat{\mathbf{W}} \bs{\beta}\|^2$. The augmented lagrangian for the above problem is:
$L(\bs{\gamma}_m,\bs{\beta},\bs{\lambda})=\frac{1}{ML}\sum_{m=1}^M \sum_{l=1}^{L} loss(\hat{W}\bm{\gamma}_m;\mathbf{x}_{ml},y_{ml}) + r(\bs{\beta})
 + \sum_{i=1}^M\frac{\rho}{2}\|\bs{\gamma}_m - \bs{\beta}\|^2 + \sum_{i=1}^M\bm{\psi}_m^T(\bs{\gamma}_m - \bs{\beta})$, where $\bm{\psi}_{m}$ is the lagrange multiplier vector corresponding to $m^{th}$ constraint. 
Let $\mathbf{A}_m \in \vect{R}^{L \times d} = - \diag{\mathbf{y}_m} \mathbf{X}_m \hat{W} $. 
 Using results from \cite{boyd11}, the ADMM updates for solving the above problem can derived as:
\begin{align}
 & \bs{\gamma}_{m}^{k+1} := \argmin_{\bs{\gamma}} (loss(\mathbf{A}_i \bs{\gamma} ) + (\rho/2)\|\bs{\gamma}_ml - \bs{\beta}^k + \mathbf{u}_{m}^k\|_2^2) \\
 & \bs{\beta}^{k+1} := \argmin_{\bs{\beta}}(r(\bs{\beta}) + (M\rho/2)\|\bs{\beta}-\overline{\bs{\gamma}}^{k+1}-\overline{\mathbf{u}}^k\|_2^2) \\
 & \mathbf{u}_{m}^{k+1} = \mathbf{u}_{m}^{k} + \bs{\gamma}_m^{k+1} - \bs{\beta}^{k+1}.
\end{align}
where, $\mathbf{u}_m = \frac{1}{\rho}\bs{\psi}_{m}$, $\overline{\bm{\gamma}} = \frac{1}{M}\sum_{m=1}^M\bm{\gamma}_m$ and $\overline{\mathbf{u}} = \frac{1}{M}\sum_{m=1}^M\mathbf{u}_m$
and the superscript $k$ denotes the iteration counts. Algorithm \ref{algo:dspt} describes the full procedure.
{\small
\begin{algorithm}[h]
\caption{Distributed Weighted Parameter Averaging (DWPA)}
\label{algo:dspt}
\SetAlgoLined
\SetKwData{Left}{left}\SetKwData{This}{this}\SetKwData{Up}{up}
\SetKwFunction{Union}{Union}\SetKwFunction{FindCompress}{FindCompress}
\SetKwInOut{Input}{input}\SetKwInOut{Output}{output}
\Input{Partitioned datasets $\mathcal{S}_m$, SVM parameter learnt for each partition $\mathbf{\hat{w}}_m, \forall m={1,\cdots, M}$}
\Output{Optimal weight vector $\bm{\beta}$}
\BlankLine
Initialize $\bm{\beta} = \mathbf{1}, \bm{\gamma}_m=\mathbf{1}, \mathbf{u}_m=\mathbf{1}, \forall m \in \{1,\cdots,M\}$\;
\While{$k < T$}{ 
\tcc{\emph{Executed on slaves}}
\For{$m\leftarrow 1$ \KwTo $M$}{
$\bm{\gamma}_m^{k} := \argmin_{\bm{\gamma}_m}(1^T (A_m\bm{\gamma}_m + 1)_+ + \rho/2 \| \bm{\gamma}_m^{k-1} - \bm{\beta}^{k-1} - \mathbf{u}_m^{k-1} \|_2^2)$
}
\BlankLine
\tcc{\emph{Executed on master}}
$\bm{\beta}^{k} := \frac{1}{2\lambda}(\hat{W}^T\hat{W} + M\rho I_m)^{-1} M\rho(\overline{\bm{\gamma}}^{k} + \overline{\mathbf{u}}^{k-1})$ ~\label{beta:update} \\
\For{$m\leftarrow 1$ \KwTo $M$}{
$\mathbf{u}_m^{k} = \mathbf{u}_m^{k-1} + \bm{\gamma}_m^{k} - \bm{\beta}^{k}$ ~\label{u:update}
}
}
\end{algorithm}
}

A heuristic called \textit{overrelaxation} \cite{boyd11} is ofter used for improving the convergence rate of ADMM. For overrelaxation, the updates for $\bm{\beta}^{k}$ (line \ref{beta:update} and $\mathbf{u}_m^{k}$ (line \ref{u:update}) are obtained by replacing $\overline{\bm{\gamma}}^{k}$ with $\bm{\hat{\gamma}}_m^{k} = \alpha\times\bm{\gamma}_m^{k} + (1-\alpha)\times\bm{\beta}^{k-1}$, in algorithm \ref{algo:dspt}. We implemented this heuristic for both DSVM and DWPA. We call them accelarated DSVM (DSVMacc) and accelarated DWPA (DWPAacc).

\subsection{Bound on stability of WPA}
\label{sec:stability}

In this section, we derive a bound of $\mathcal{O}(\frac{1}{ML})$ on stability of the final hypothesis returned by WPA algorithm described in eqn. \ref{eq:nondistopt}. A similar bound was derived by Mann et al. \cite{mann09} on the stability of PA. This leads to a $\mathcal{O}(\frac{1}{\sqrt{ML}})$ bound on deviation from optimizer of generalization error (see \cite{mann09}, theorem 2).


Let $S = \{S_1,\cdots,S_M\}$ and $S' = \{S'_1,\cdots,S'_M\}$ be two datasets with $M$ partitions and $L$ datapoints per partition, differing in only one datapoint. Hence, $S_m = \{z_{m1},\cdots,z_{mL}\}$ and $S'_m = \{z'_{m1},\cdots,z'_{mL}\}$, where $z_{ml} = (\mathbf{x}_{ml},y_{ml})$ and  $z'_{ml} = (\mathbf{x}'_{ml},y'_{ml})$. Further, $S_1 = S'_1,\cdots,S_{M-1} = S'_{M-1}$, and $S_M$ and $S'_M$ differs at single point $z_{ML}$ and $z'_{ML}$. Also, let $\|\mathbf{x}\| \leq R$, $\forall \mathbf{x}$.
%
Moreover, let $\hat{W} = [\hat{\mathbf{w}}_{S_1},\cdots,\hat{\mathbf{w}}_{S_M}]$ and $\hat{W'} = [\hat{\mathbf{w}}_{S'_1},\cdots,\hat{\mathbf{w}}_{S'_M}]$ where,
$  \hat{\mathbf{w}}_{S_i} =  \argmin_\mathbf{w} \lambda\|\mathbf{w}\|^2 + \frac{1}{L}\sum_{i \in S_i} max(0,1-y\mathbf{w}^T\mathbf{x})$.
We also assume $\|\hat{\mathbf{W}}\|_{F} = \|\hat{\mathbf{W}}'\|_{F} = 1 $. Hence, $\|\hat{\mathbf{w}}_m\|^2 = \|\hat{\mathbf{w}}'_m\|^2 = \frac{1}{M}, \forall m \in \{1,\cdots,M\}$. 


We also define the following quantities:
{\small
 \[ \bm{\beta} = \argmin_{\bm{\beta}} \lambda\|\hat{W}\bm{\beta}\|^2 + \frac{1}{ML}\sum_{i=1}^M \sum_{z \in S_i} max(0,1-y(\hat{W}\bm{\beta})^T\mathbf{x}) \]
 \[ \bm{\beta'} = \argmin_{\bm{\beta}} \lambda \|\hat{W'}\bm{\beta}\|^2 + \frac{1}{ML}\sum_{i=1}^M \sum_{z' \in S'_i} max(0,1-y'(\hat{W}\bm{\beta})^T\mathbf{x'}) \]
 \[ \bm{\tilde{\beta}} = \argmin_{\bm{\beta}} \lambda \|\hat{W'}\bm{\beta}\|^2 + \frac{1}{ML}\sum_{i=1}^M \sum_{z \in S_i} max(0,1-y(\hat{W'}\bm{\beta})^T\mathbf{x}) \]
}
Also, let $\bm{\theta} = \hat{W}\bm{\beta}$, $\bm{\theta}' = \hat{W'}\bm{\beta}'$ and $\tilde{\bm{\theta}} = \hat{W'}\tilde{\bm{\beta}}$.

We are interested in deriving a bound on $\|\bm{\theta} - \bm{\theta'}\|$, which decompose as:
$\|\bm{\theta} - \bm{\theta'}\| \leq \|\bm{\theta} - \tilde{\bm{\theta}}\| + \|\tilde{\bm{\theta}} - \bm{\theta'}\|$. Intuitively, the first term captures the change from $\hat{W}$ to $\hat{W}'$ and second term captures change in dataset.
Lemma ~\ref{lemma1}, shows that $\|\tilde{\bm{\theta}} - \bm{\theta'}\|$ is $\mathcal{O}(\frac{1}{ML})$.
Showing bound on $\|\bm{\theta} - \tilde{\bm{\theta}}\|$ requires bounds on $\|\bm{\beta} - \tilde{\bm{\beta}}\|$ (lemma \ref{lemma2}) and $\|\hat{W}-\hat{W}'\|$ (lemma \ref{theorem1}). The final proof is given in Theorem \ref{theorem:bound}.


\begin{lem}
 \label{theorem1}
$\|\hat{W}-\hat{W}'\| = \mathcal{O}(\frac{1}{ML})$
\end{lem}
\textbf{Proof (sketch): } 
Since  $\hat{w}_m=\hat{w}'_m, m=1,\dots,M-1$, it suffices to show that $\|\hat{w}_M - \hat{w}'_M\| = \mathcal{O}(\frac{1}{ML})$. Since, $\hat{w}$ and $\hat{w}'$ are scaled as $\|\hat{w}_m\|^2=\|\hat{w}'_m\|^2=\frac{1}{M}$ it suffices to show that $M\|\hat{w}-\hat{w}'\| = \mathcal{O}(\frac{1}{L})$. This result is analogous to theorem 1 of \cite{mann09}. This can be proved using a special definition of bregmann divergence shown in appendix A.

\begin{lem}
\label{lemma1}
$\|\tilde{\bm{\theta}} - \bm{\theta'}\| = \mathcal{O}(\frac{1}{ML})$
\end{lem}

\textbf{Proof (sketch):} This can be shown using similar technique as proof in appendix B using $\|\cdot\|_K$, where, $K = \hat{W'}^T\hat{W'}$ instead of the Euclidean norm.

\begin{lem}
\label{lemma2} 
$\|\bm{\beta} - \tilde{\bm{\beta}}\| = \mathcal{O}(\frac{1}{ML})$
\end{lem}
\textbf{Proof: } 
Let $F_W(\bm{\beta}) = G_W(\bm{\beta}) + L_W(\bm{\beta})$ and $F_{W'}(\tilde{\bm{\beta}}) = G_{W'}(\tilde{\bm{\beta}}) + L_{W'}(\tilde{\bm{\beta}})$.
Using a similar definition of Bregmann divergence as in appendix B and its positivity:
\begin{equation}
\label{bregineq1}
 B_{G_{\hat{W}}}(\tilde{\bm{\beta}}\|\bm{\beta}) + B_{G_{\hat{W'}}}(\bm{\beta}\|\tilde{\bm{\beta}}) \leq B_{F_{\hat{W}}}(\tilde{\bm{\beta}}\|\bm{\beta}) + B_{F_{\hat{W'}}}(\bm{\beta}\|\tilde{\bm{\beta}})
\end{equation}

The \textit{left hand side} of the inequality ~\ref{bregineq1}, is given by;
\begin{align}
 B_{G_{\hat{W}}}(\tilde{\bm{\beta}}\|\bm{\beta}) + B_{G_{\hat{W'}}}(\bm{\beta}\|\tilde{\bm{\beta}}) &= \lambda\|\bm{\tilde{\beta}} - \bm{\beta}\|^T (\|\hat{W}^T\hat{W} + \hat{W'}^T\hat{W'}\|)\|\bm{\tilde{\beta}} - \bm{\beta}\|\nonumber\\ 
 &\leq \lambda\|\bm{\beta} - \tilde{\bm{\beta}}\|_{K'}^2,\mbox{ where, }K'=\hat{W}^T\hat{W} + \hat{W'}^T\hat{W'}\nonumber
\end{align}

Now we solve the \textit{right hand side} of inequality ~\ref{bregineq1},
\begin{align}
~\label{bregfunc1}
 B_{F_{\hat{W}}}(\tilde{\bm{\beta}}\|\bm{\beta}) + B_{F_{\hat{W'}}}(\bm{\beta}\|\tilde{\bm{\beta}}) &= F_{\hat{W}}(\bm{\tilde{\beta}}) - F_{\hat{W}}(\bm{\beta}) + F_{\hat{W'}}(\bm{\beta}) + F_{\hat{W'}}(\bm{\tilde{\beta}} )\nonumber\\
 =& \lambda[\|\hat{W}\bm{\tilde{\beta}}\|^2 - \|\hat{W}\bm{\beta}\|^2 + \|\hat{W'}\bm{\beta}\|^2 - \|\hat{W'}\bm{\tilde{\beta}}\|^2] + \\
 &[L_{\hat{W}}(\bm{\beta}') - L_{\hat{W}}(\bm{\beta}) + L_{\hat{W'}}(\bm{\beta}) - L_{\hat{W'}}(\bm{\beta}')] = \mathcal{R} + \mathcal{L} \nonumber  
\end{align}
From ~\ref{bregfunc1}, we have,
{\small
\begin{align}
 \mathcal{L} &= L_{\hat{W}}(\bm{\beta}') - L_{\hat{W}}(\bm{\beta}) + L_{\hat{W'}}(\bm{\beta}) - L_{\hat{W'}}(\bm{\beta}')\nonumber\\
 &= \frac{1}{ML}\sum_{m,l=1}^{M,L}[max(0,1-y_{ml}(\hat{W}\bm{\tilde{\beta}})^T\mathbf{x}_{ml}) - max(0,1-y_{ml}(\hat{W}\bm{\beta})^T\mathbf{x}_{ml}) + \nonumber \\
 & max(0,1-y_{ml}(\hat{W'}\bm{\beta})^T\mathbf{x}_{ml}) - max(0,1-y_{ml}(\hat{W'}\bm{\tilde{\beta}})^T\mathbf{x}_{ml})]\nonumber\\
 &\leq \frac{1}{ML}\sum_{m,l=1}^{M,L}max(0,y_{ml}((\hat{W'} - \hat{W})(\bm{\beta} - \bm{\tilde{\beta}}))^T\mathbf{x}_{ml}) \leq \frac{1}{ML}\sum_{m,l=1}^{M,L}|y_{ml}((\hat{W'} - \hat{W})(\bm{\beta} - \bm{\tilde{\beta}}))^T\mathbf{x}_{ml}|\nonumber\\
 &\leq \frac{R}{ML}\|(\bm{\beta} - \bm{\tilde{\beta}}))\|\nonumber
\end{align}
}
Where, first two inequalities use: $max(a,0) - max(b,0) \leq max(a-b,0)$, and the last step uses lemma \ref{theorem1}. 

For the part $\mathcal{R}$ of the ~\ref{bregfunc1} involving regularizers:
\begin{align}
 \mathcal{R} &=  \lambda[\|\hat{W}\bm{\tilde{\beta}}\|^2 - \|\hat{W}\bm{\beta}\|^2 + \|\hat{W'}\bm{\beta}\|^2 - \|\hat{W'}\bm{\tilde{\beta}}\|^2]
 =\lambda(\tilde{\bm{\beta}}+\bm{\beta})^T(\hat{W}^T\hat{W} - \hat{W'}^T\hat{W'})(\tilde{\bm{\beta}} - \bm{\beta})\nonumber\\
 &\leq \lambda \|\bm{\tilde{\beta}} + \bm{\beta}\| \| \hat{W} \| \|(\hat{W} - \hat{W'})\| \|\bm{\tilde{\beta}} - \bm{\beta}\| + \|\bm{\tilde{\beta}} + \bm{\beta}\| \| \hat{W'}\| \|(\hat{W} - \hat{W'})\| \|\bm{\tilde{\beta}} - \bm{\beta}\| \nonumber \\
 &\leq \frac{4\lambda R}{ML}  \|\bm{\beta} - \bm{\tilde{\beta}}\|\nonumber
\end{align}
where for the last step, we use the constant bound $\|\bm{\beta}\| = \lambda R$ on $\bm{\beta}$ obtained from its expression of in ~\ref{beta:dual}.
Therefore, from the \textit{left hand side} and \textit{right hand side} of the inequality ~\ref{bregineq1}, we have:
\begin{equation}
 \lambda\|\bm{\beta} - \tilde{\bm{\beta}}\|_2^2 \leq \frac{\lambda}{\sigma_{min}}\|\bm{\beta} - \tilde{\bm{\beta}}\|_{K'}^2 \leq \frac{4\lambda R}{ML}  \|\bm{\beta} - \bm{\tilde{\beta}}\| + \frac{R}{ML}\|(\bm{\beta} - \bm{\tilde{\beta}}))\|
\end{equation}
 where, $\sigma_{min}$ is smallest eigenvalue of $K'$. This implies;$\|\bm{\beta} - \tilde{\bm{\beta}}\|$ is $\mathcal{O}(\frac{1}{ML})$.

\qed

\begin{thm}
 \label{theorem:bound}
$\|\bm{\theta} - \bm{\theta'}\|$ is of the order of $\mathcal{O}(\frac{1}{ML})$.
\end{thm}
\textbf{Proof: } The steps involved in the proof are as follows;
\begin{equation}
 \|\bm{\theta} - \bm{\theta'}\| \leq \|\bm{\theta} - \tilde{\bm{\theta}}\| + \|\tilde{\bm{\theta}} - \bm{\theta'}\|
\end{equation}
From lemma ~\ref{lemma1}, $\|\|\bm{\theta} - \tilde{\bm{\theta}}\|$ is $\mathcal{O}(\frac{1}{ML})$.

\begin{align}
~\label{final:proof}
 \|\tilde{\bm{\theta}} - \bm{\theta'}\| &\leq \frac{1}{2}(\|(\hat{W}-\hat{W'})(\bm{\beta} + \tilde{\bm{\beta}})\| + \|(\hat{W}+\hat{W'})(\bm{\beta} - \tilde{\bm{\beta}})\|)\nonumber\\
 &\leq \frac{1}{2}((\|(\hat{W}-\hat{W'}\|)(\|\bm{\beta} + \tilde{\bm{\beta}}\|) + (\|\hat{W}+\hat{W'}\|)(\|\bm{\beta} - \tilde{\bm{\beta}}\|))
\end{align}
We have already, shown that, we have a constant bound on $\|\bm{\beta} + \tilde{\bm{\beta}}\|$ and $\|\hat{W}+\hat{W'}\|_F$, since, norms of $\bm{\beta}$, $\bm{\tilde{\beta}}$, 
$\hat{W}$ and $\hat{W'}$ are bounded.
Also both $\|\bm{\beta} - \tilde{\bm{\beta}}\|$ 
and $\|(\hat{W}-\hat{W'}\|$ are $\mathcal{O}(\frac{1}{ML})$.

Hence, from ~\ref{final:proof}, we have the required result.

\qed


\section{Experimental Results}
\label{sec:exp}
\vspace{-2mm}

In this section, we experimentally analyze and compare the methods proposed here, \textit{distributed weighted parameter averaging} (DWPA) and accelerated DWPA (DWPAacc) described in section \ref{sec:distalgo}, with \textit{parameter averaging} (PA) \cite{mann09}, Distributed SVM (DSVM) using ADMM, and accelerated DSVM (DSVMacc) \cite{boyd11}.
For our experimentation, we have implemented all the above mentioned algorithms in Matlab \cite{MATLAB:2010}. We have used the \textit{liblinear} library \cite{REF08a} to obtain the SVM parameters corresponding each partition. Optimization problems which arise as subproblems in ADMM has been solved using CVX \cite{gb08}, \cite{cvx}.

We used both toy datasets (section \ref{sec:toy}) and real world datasets (described in table \ref{datasets}) for our experiments. Real world datasets were obtained from LIBSVM website \cite{libsvmdata}. Samples for real world datasets were selected randomly. The datasets were selected to have various ranges of feature count and sparsity. Section \ref{sec:toy} describes a specially construc

{\small
 \begin{table}[!htbp]
  \caption{\label{datasets} Training and test dataset size} 
  \vspace{-2mm}
 \centering
 \small
 \begin{tabular}{|c|c|c|c|c|}
 \hline
  Dataset Name& Number of training instances & Number of test instances& Number of features&Domain\\\cline{1-5}
  \hline
  epsilon& 6000 & 1000&	2000&mixed\\\cline{1-5} 
  \hline
  gisette&6000 & 1000&	5000&mixed \\\cline{1-5}
  \hline
  real-sim& 3000& 5000&	20958&text\\\cline{1-5}
  \hline
 \end{tabular}
 \end{table}
}


\subsection{Results on toy dataset}
\label{sec:toy}
\vspace{-2mm}

\begin{figure}[!ht]
 \captionsetup[subfigure]{labelformat=empty}
  \subfloat[]{
  \includegraphics[width=0.33\linewidth]{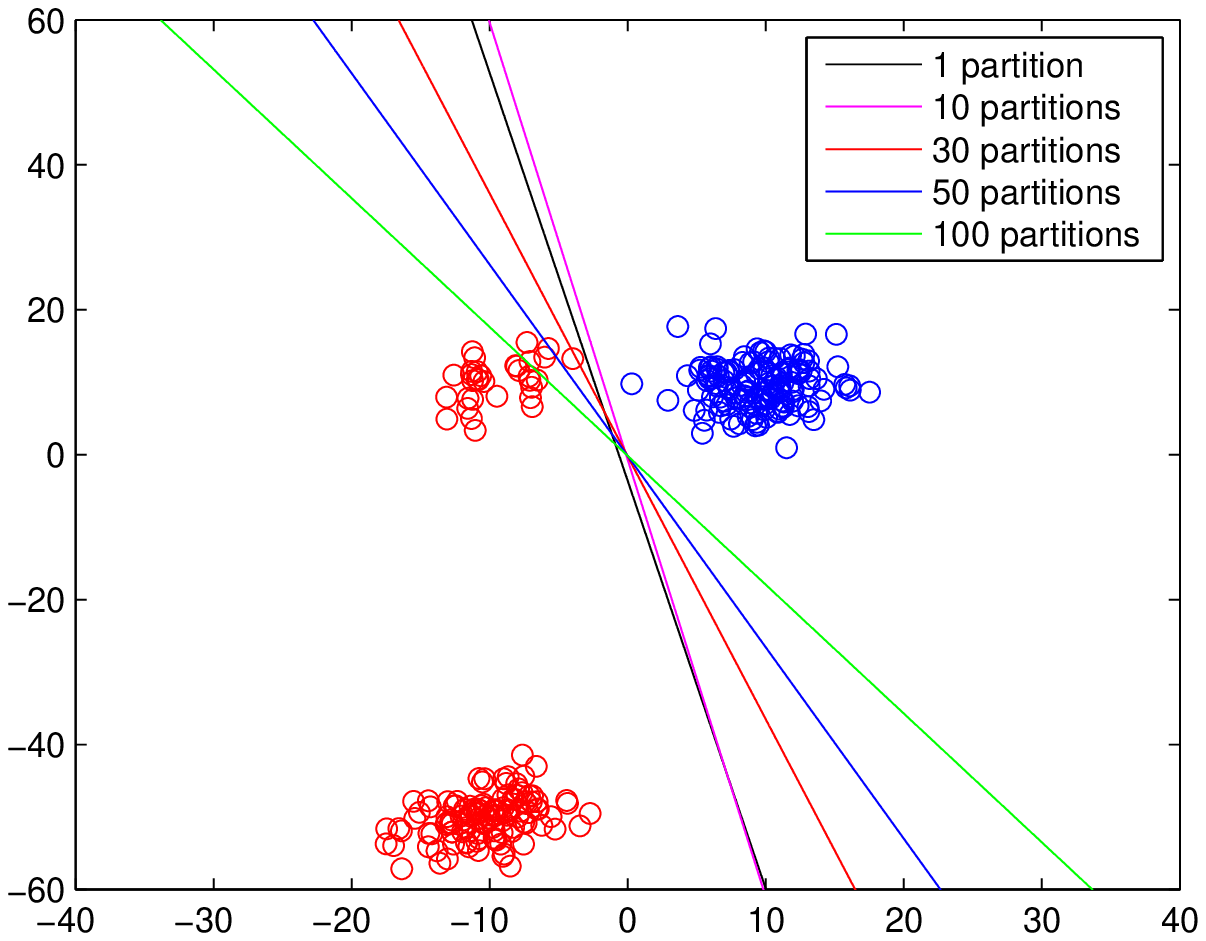}
  }
  \hspace{-.5cm}
  \subfloat[]{
  \includegraphics[width=0.33\linewidth]{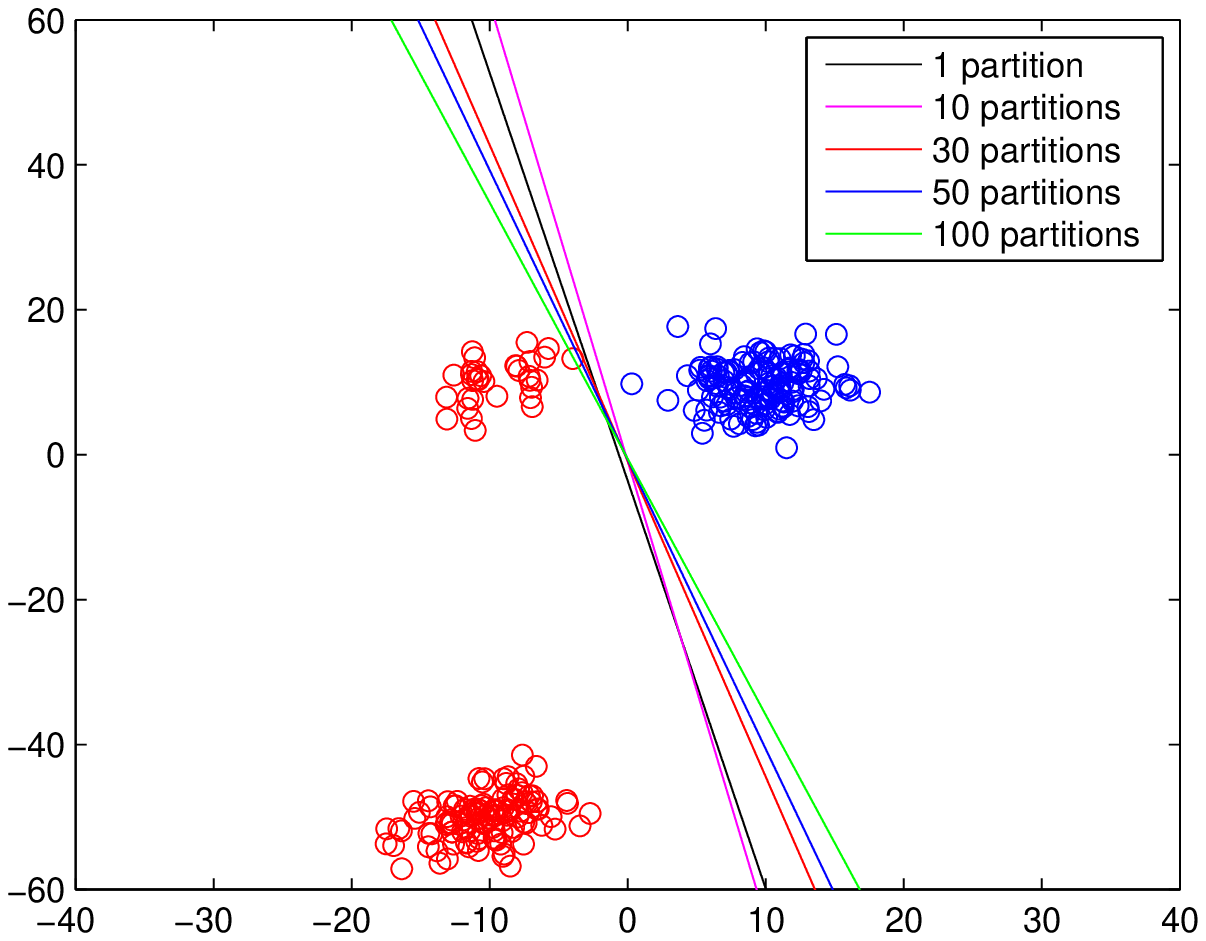}
  }
  \hspace{-.5cm}
  \subfloat[]{
  \includegraphics[width=0.33\linewidth]{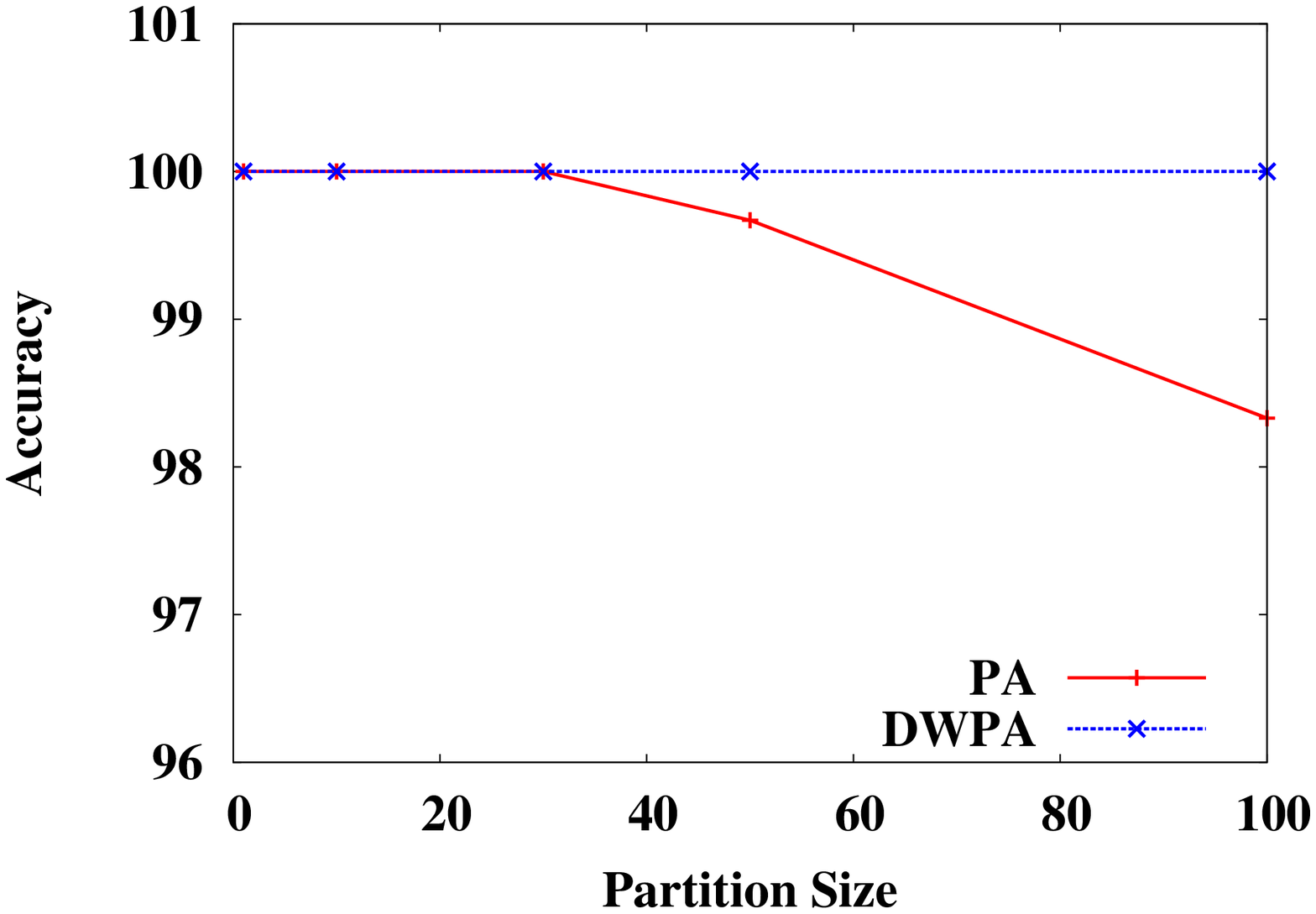}
  }
  \vspace{-7mm}
  \caption{\label{toy} Comparison of performance of {\it PA}(left) and {\it DWPA}(middle) on toy dataset. The graph on the \textit{right} shows the 
  change in accuracy of PA and DWPA with change in partition size for the toy dataset}
 \end{figure}

The main purpose of the toy dataset was to visually observe the effect of change in the number of partitions on the final hypothesis for various algorithms. Datapoints are generated from a 2 dimensional mixture of gaussians.
In figure \ref{toy}, the \red{red} and \blue{blue} dots indicate the datapoints from two different classes. The upper red blob contains only 20\% of red points. Hence as the number of partitions increase, many partitions will not have any  data points from upper blob. For these partitions, the separating hyperplane passes throught the upper red blob. These cause the average hyperplane to pass through upper red blob, thus decreasing the accuracy. This effect is visible in the left plot of figure \ref{toy}. This effect is mitigated in weighted parameter averaging as the weights learnt for the hyperplanes passing through upper red blob are lesser. This is shown in middle plot of figure \ref{toy}. Finally, the right plot of figure \ref{toy} shows the resultant decrease in accuracy for PA with increase in number of partitions.

Bias of a learning algorithm is $E[|w - w*|]$, where $w$ and $w*$ are minimizers of regularized loss and generalization error, and the expectation is over all samples of fixed size say $N$. An criticism against PA is the lack of bound on bias \cite{mann09}. In table \ref{biascomparison}, we compare bias of PA, WPA and SVM as a function of $N$. Data samples were generated from the same toy distribution as above. $w*$ was computed by training on a large sample size and ensuring that training set error and test set error are very close. $w$ was computed 100 times by randomly sampling from the distribution. The average of $|w-w*|$ is reported in table \ref{biascomparison}. We observe that bias of PA is indeed much higher than SVM or WPA.

  \begin{table}[!ht]
   \caption{\label{biascomparison} Variation of mean bias with increase in dataset size for PA, DWPA and DSVM}
   \vspace{-2mm}
  \centering
  \small
  \begin{tabular}{|c|c|c|c|}
	\hline
	Sample size &Mean bias(PA)&Mean bias(DWPA)&Mean bias(DSVM)\\\cline{1-4}
	3000 	&0.868332	&0.260716	&0.307931	\\\cline{1-4}
	6000 	&0.807217	&0.063649	&0.168727	\\\cline{1-4}
	\hline
  \end{tabular}
  \end{table}

\subsection{Comparison of Accuracies}
\label{sec:acc}
\vspace{-2mm}

\begin{figure}[!ht]
  \captionsetup[subfigure]{labelformat=empty}
  \begin{center}
  \subfloat[]{
   \includegraphics[width=0.33\linewidth]{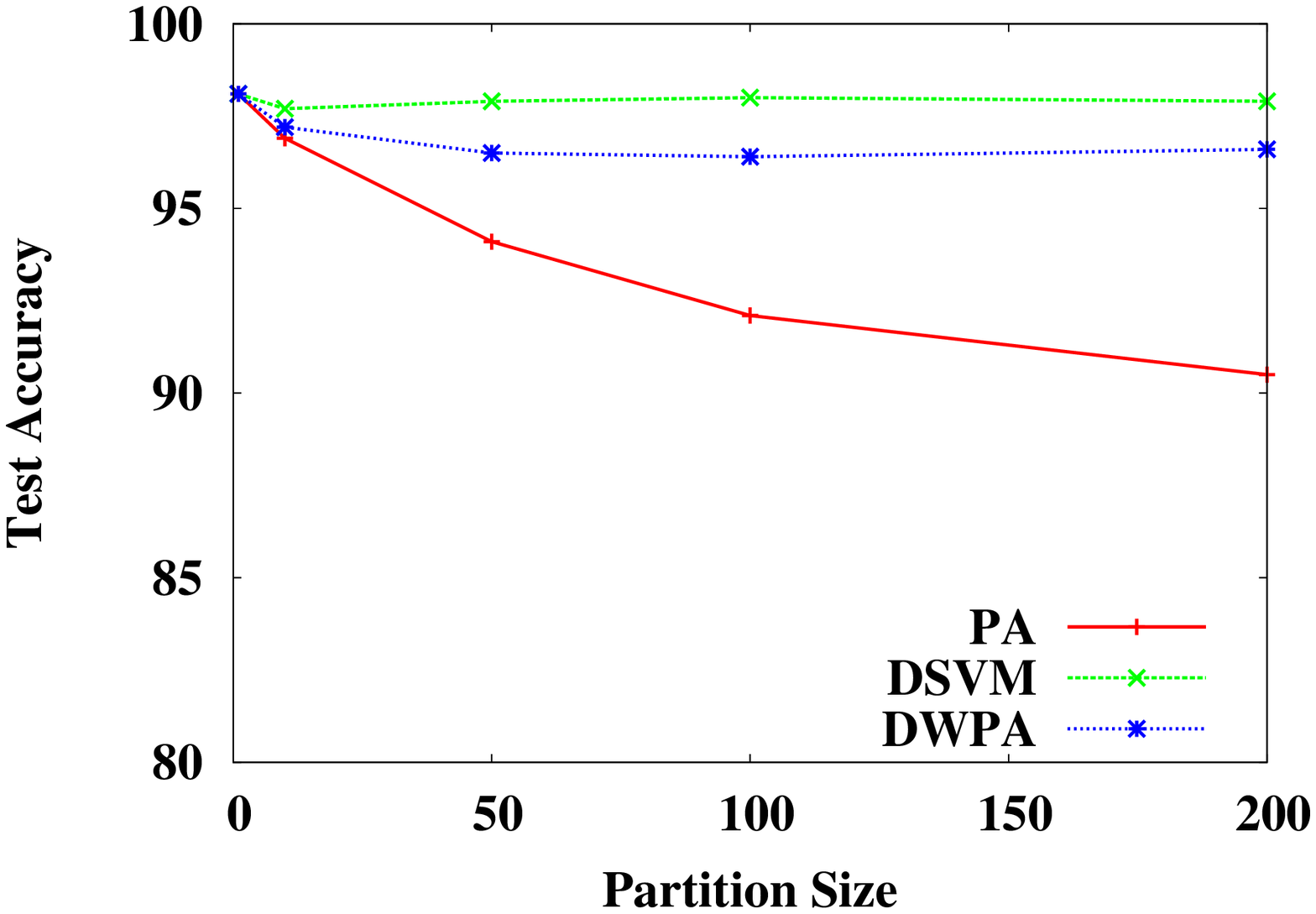}
  }
  \hspace{-.5cm}
  \subfloat[]{
  \includegraphics[width=0.33\linewidth]{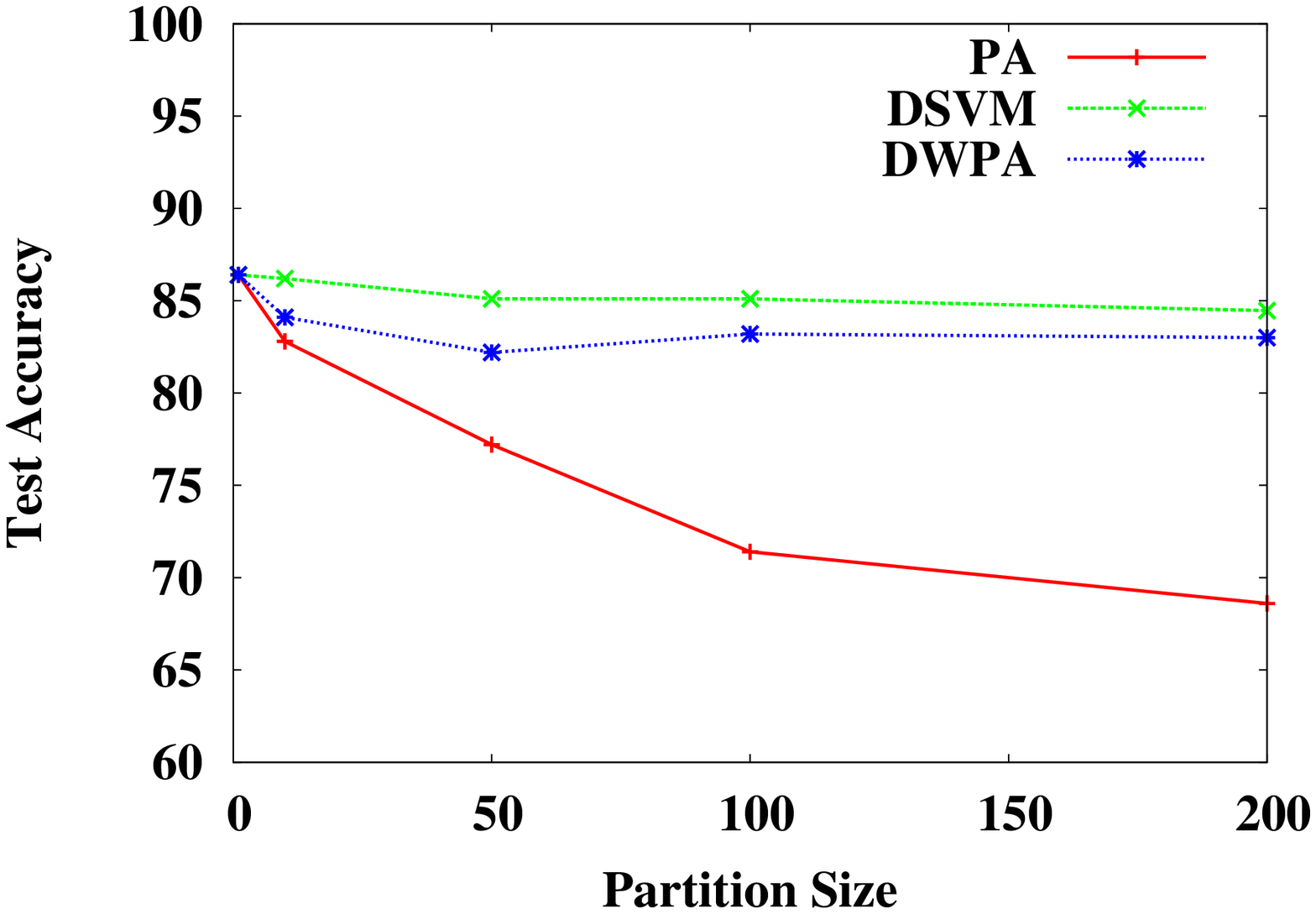}
  }
  \hspace{-.5cm}
  \subfloat[]{
  \includegraphics[width=0.33\linewidth]{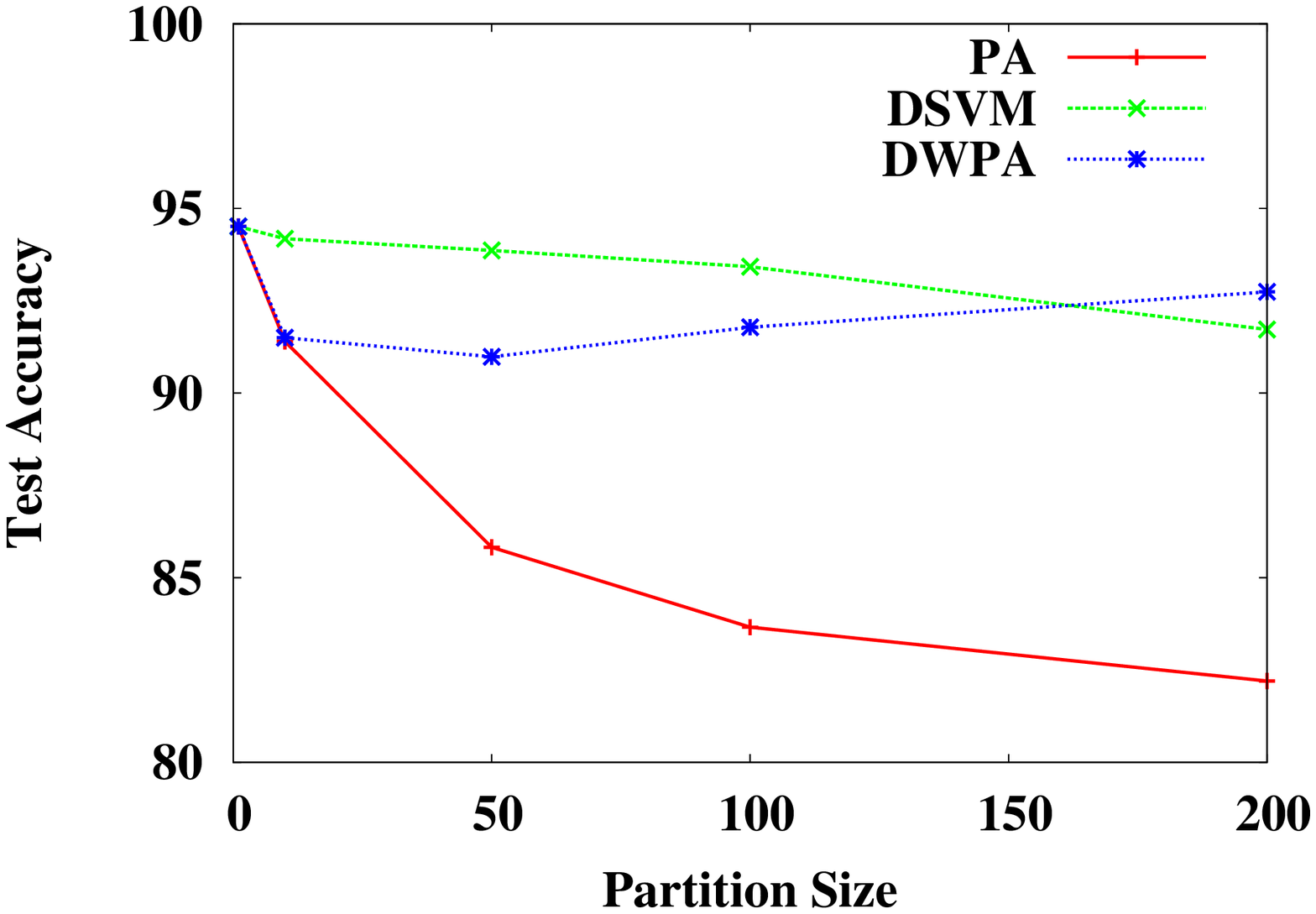}
  }
  \end{center}
  \vspace{-7mm}
  \caption{\label{ushape} Variation of accuracy with number of partitions for \textit{gisette(left), epsilon(middle)} and \textit{real-sim(right)} for
   partition size 1, 10, 50, 100 and 200. The results were recorded upto 500 iterations}
 \end{figure}

In this section, we compare accuracies obtained by various algorithms on real world datasets, with increase in number of paritions. Figure \ref{ushape} reports test set accuracies for PA, WPA and SVM on three real world datasets with varying size of partitions. It is clear that performance of PA degrades dramatically as the number of parition increases. Thus, the effect demonstrated in section \ref{sec:toy} is also observed on real world datasets.

We also observe that performance of WPA improves with increase in number of paritions. This is due to fact that dimension of space on which $\bm{x}_{ml}$'s are projected using $\mathcal{H}$ (section \ref{sec:formulation}) increases, thus reducing the information loss caused by projection. Finally, as expected WPA performs slightly worse than SVM.

\subsection{Convergence Analysis and time comparison}
\label{sec:conv}
\vspace{-2mm}

In this section, we compare the convergence properties of DSVM, DSVMacc, DWPA, and DWPAacc. Here we report results on real-sim due to lack of space. Results on other real world datasets are provided in appendix C. Top row of figure \ref{fig:conv} shows variation of primal residual (disagreement between parameters on various partitions) with iterations. It is clear that DWPA and DWPAacc show much lesser disagreement compared to DSVM and DSVMacc, thus showing faster convergence. Bottom row fo figure \ref{fig:conv} shows variation of test set accuracy with iterations. The same behaviour is apparent here, with testset accuracy of DWPA and DWPAacc converging much faster than DSVM and DSVMacc. One of the reasons is also that DWPA has an obvious good starting point of $\bm{\beta}=[\frac{1}{M},\dots,\frac{1}{M}]$ corresponding to PA.

  \begin{figure}[!ht] 
   \captionsetup[subfigure]{labelformat=empty}
  \begin{center}
   \subfloat[]{
   \includegraphics[width=0.34\linewidth]{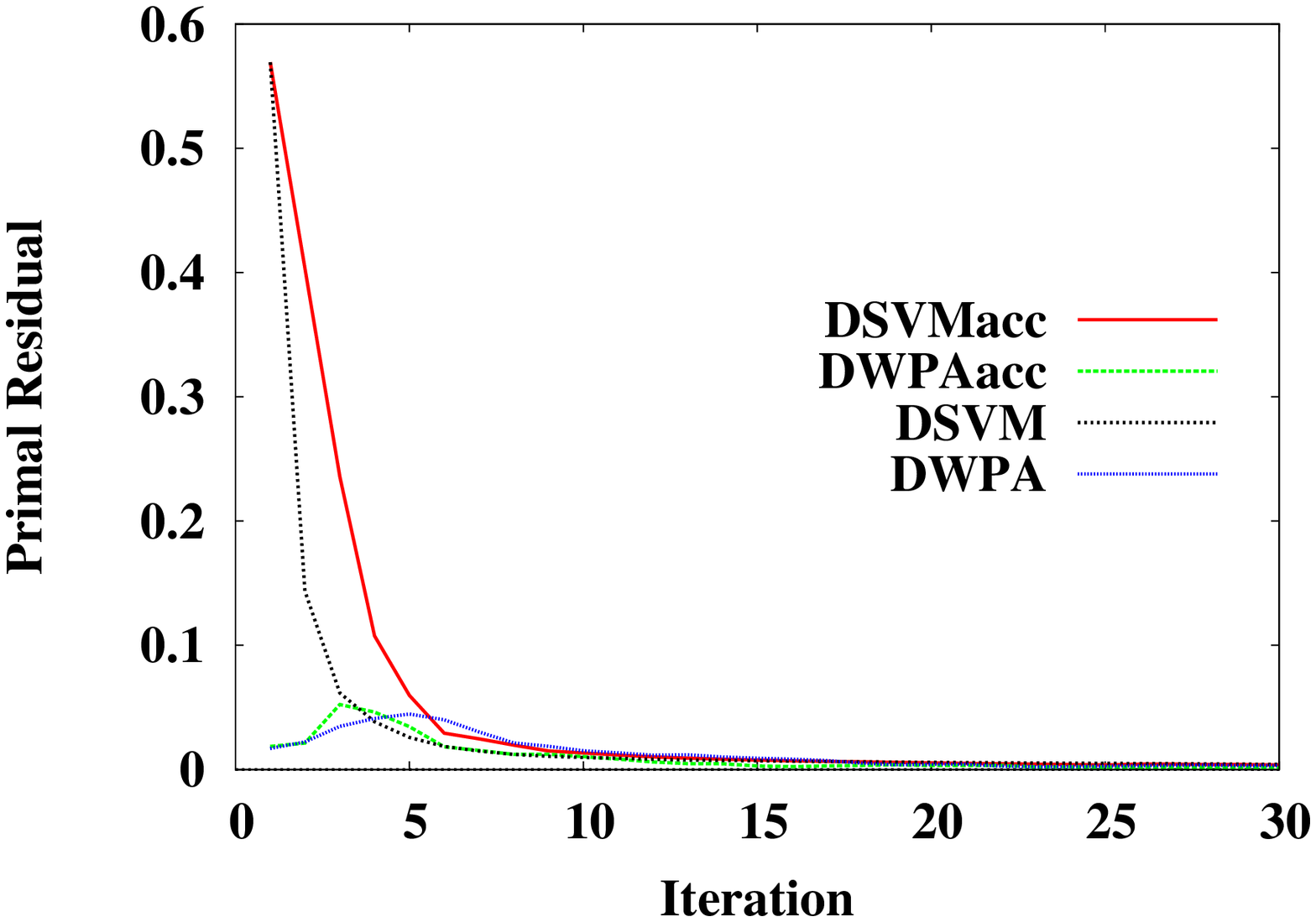}
    }
   \hspace{-.45cm}
    \subfloat[]{
   \includegraphics[width=0.34\linewidth]{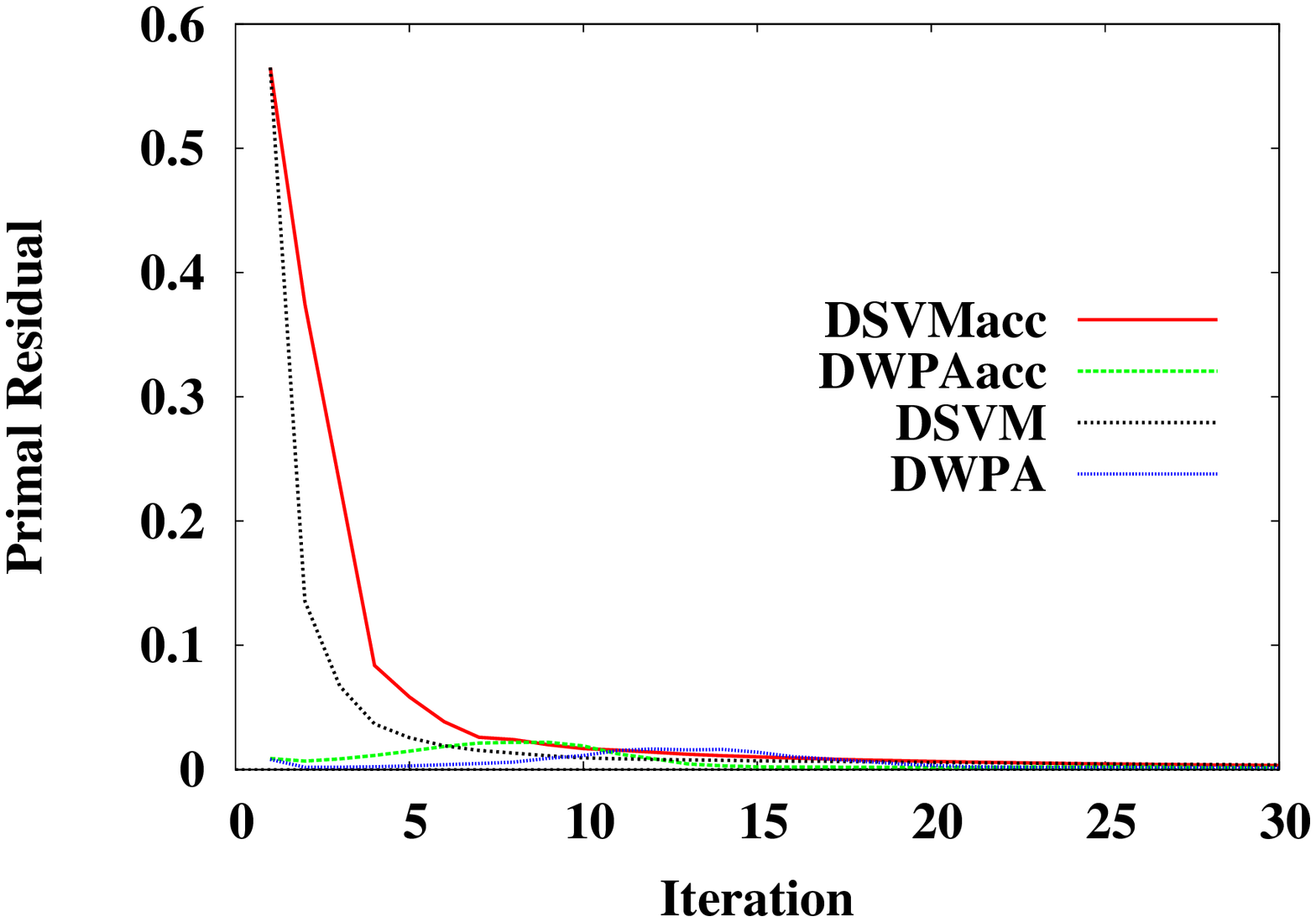}
    }
   \hspace{-.45cm}
    \subfloat[]{
   \includegraphics[width=0.34\linewidth]{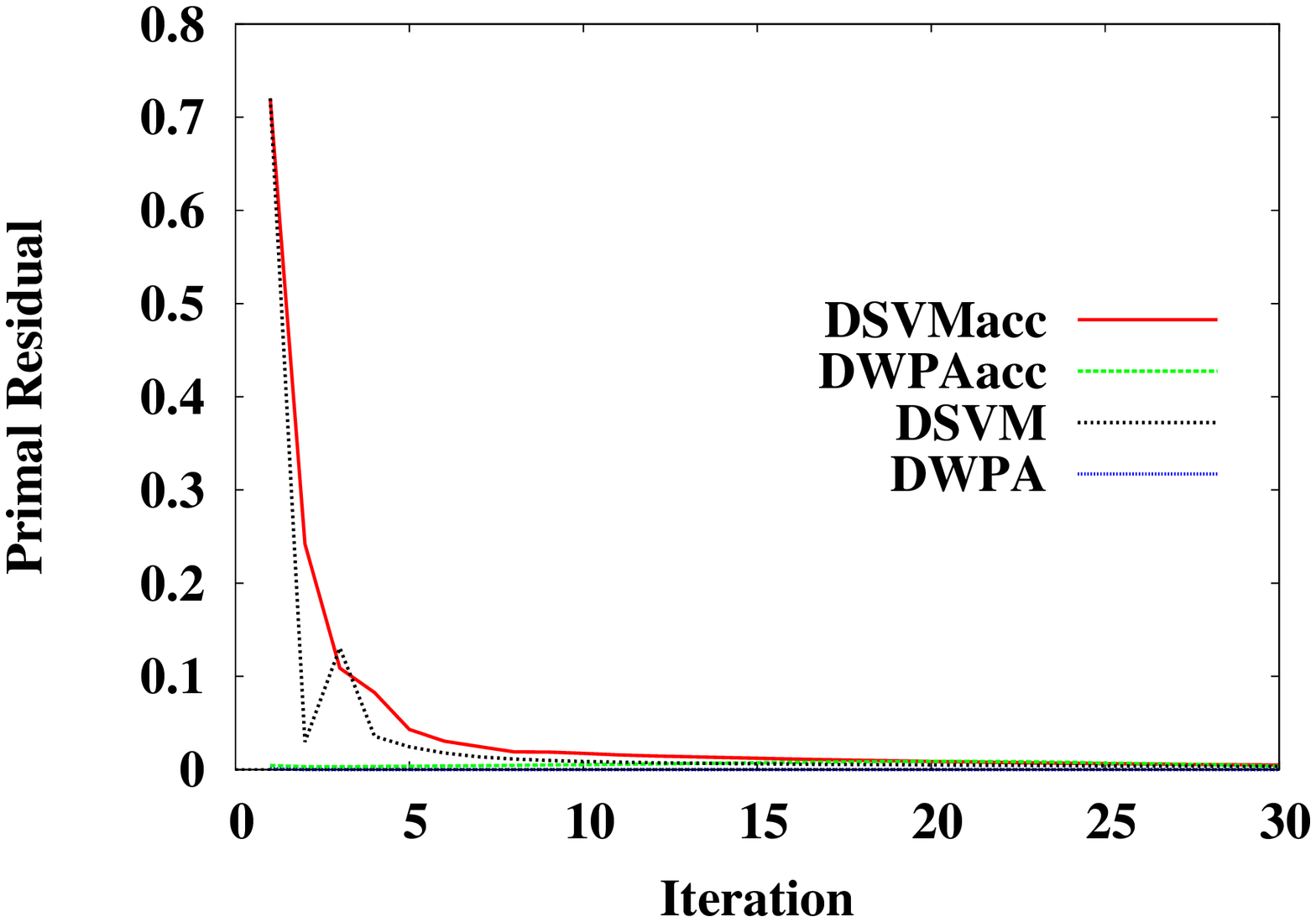}
    }
    \hspace{-.45cm}
   \subfloat[]{
   \includegraphics[width=0.34\linewidth]{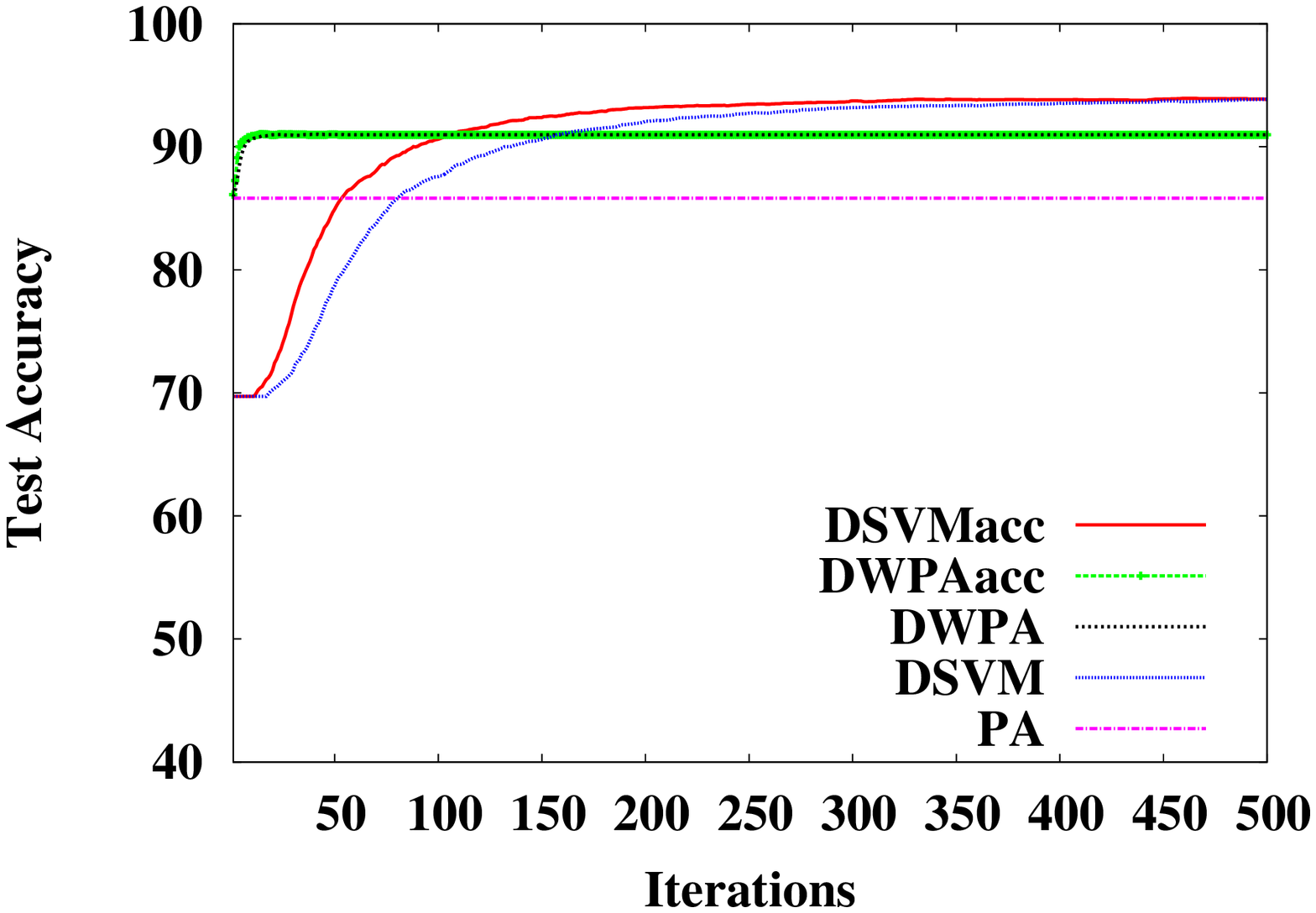}
    }
   \hspace{-.45cm}
    \subfloat[]{
   \includegraphics[width=0.34\linewidth]{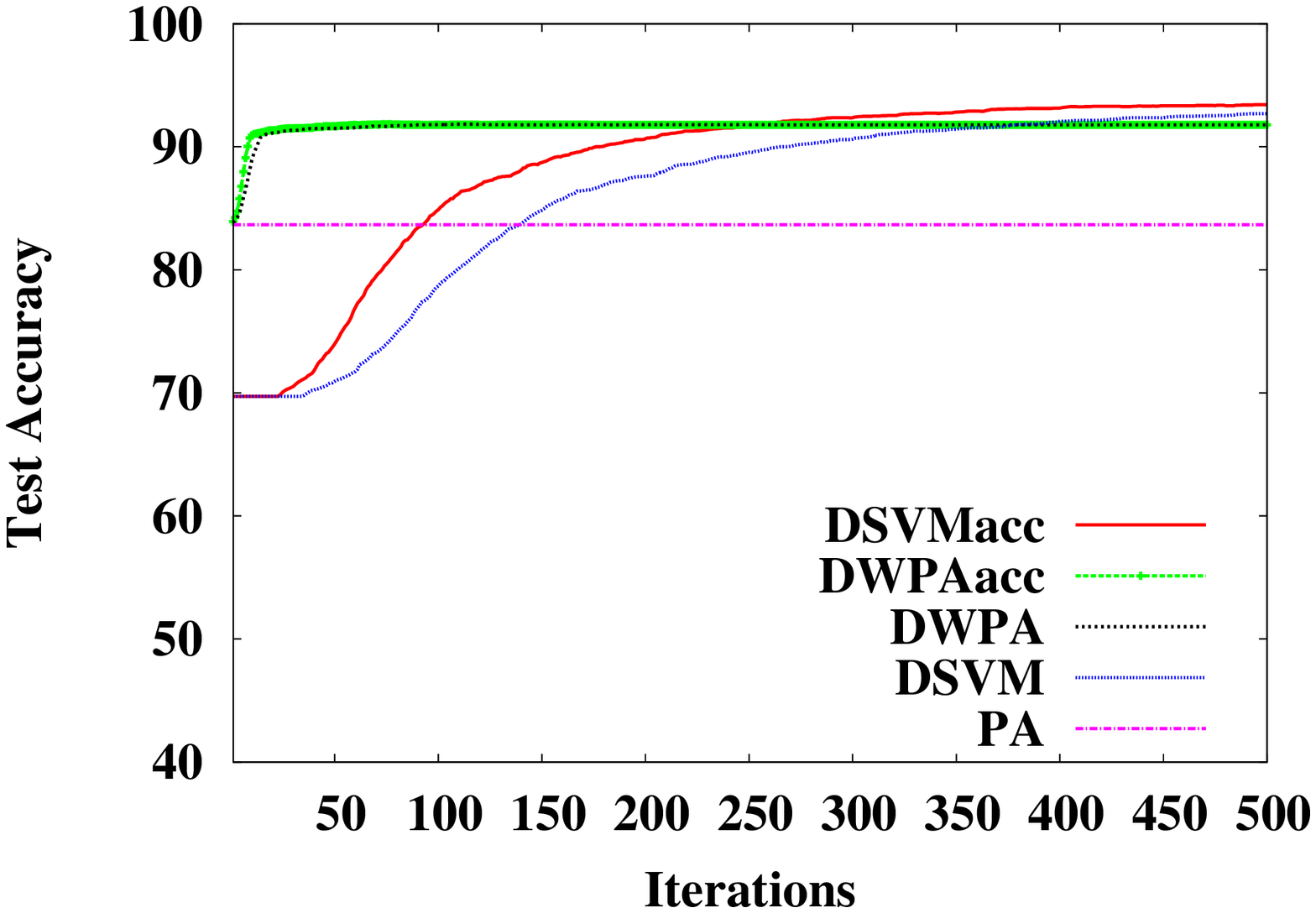}
    }
    \hspace{-.45cm}
    \subfloat[]{
   \includegraphics[width=0.34\linewidth]{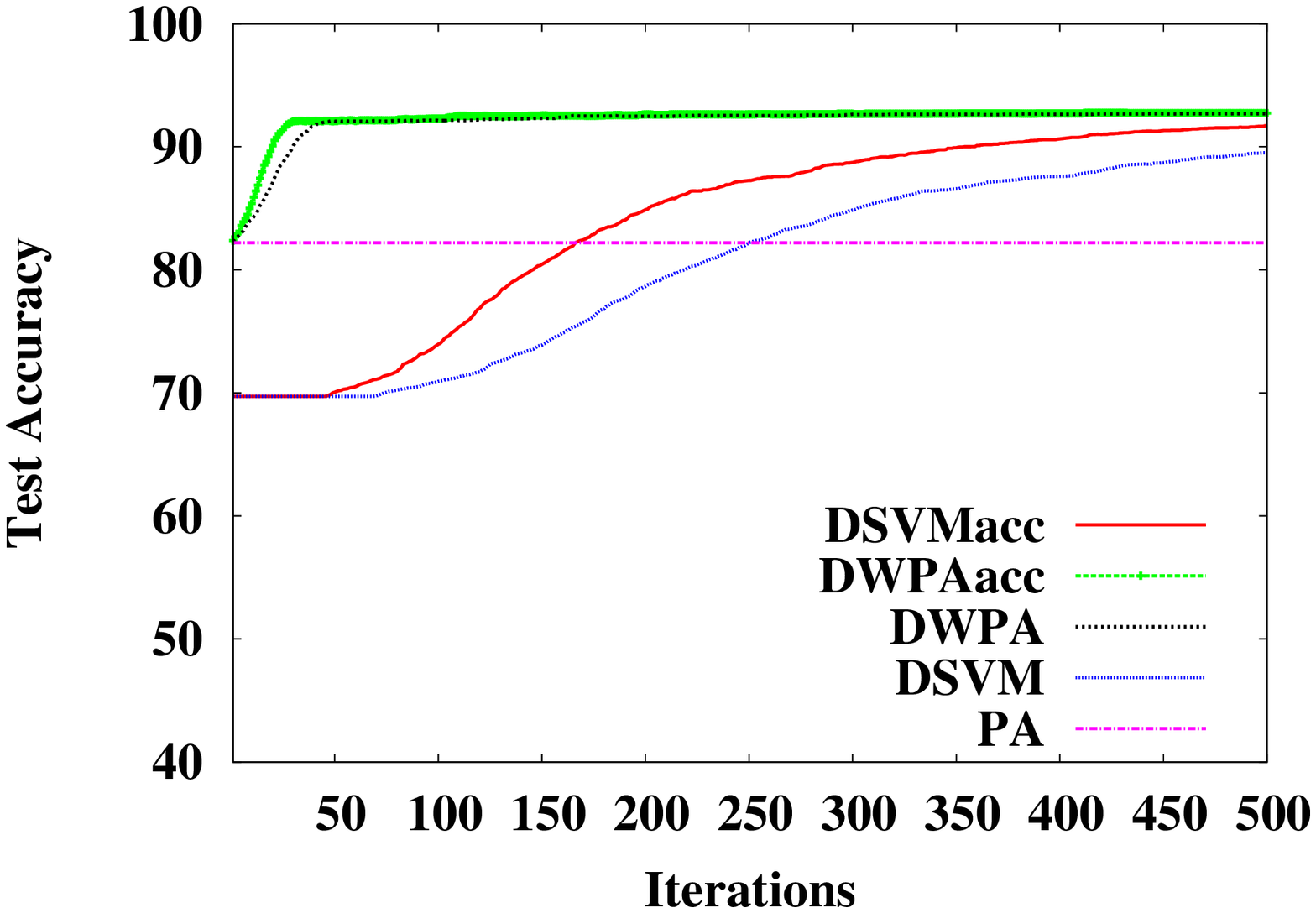}
    }

 \end{center}
   \caption{Convergence of primal residual (top) and test accuracy (bottom) for \textbf{real-sim}}\label{fig:conv}
 \end{figure}

 

\begin{center}
 \begin{table*}[!ht]
  \caption{\label{timeperiter} Average time per iteration(in seconds)} 
  \vspace{-2mm}
 \centering
 \begin{tabular}{|c|c|c|c|c|c|c|}
	\hline
	\multicolumn{1}{|c|}{} & \multicolumn{3}{|c|}{DWPA}  & \multicolumn{3}{|c|}{DSVM} \\\cline{1-7}
	\hline
	\specialcell[b]{Number of\\partitions}&epsilon & real-sim & gisette& epsilon& real-sim& gisette \\\cline{1-7}
	10   &27.4313	&13.9004	&31.2253	&624.0198	&622.6011	&1653.0 \\\cline{1-7}
	50   &23.1451	&23.1181	&37.3698	&125.0944	&125.0944	&525.7135 \\\cline{1-7}
	100  &37.3016	&47.4931	&65.1963	&116.8604	&116.8604	&440.6123 \\\cline{1-7}
	\hline
	 \end{tabular}
 \end{table*}
\end{center}

Table \ref{timeperiter} reports the average time taken by DWPA and DSVM for completing one iteration as a function of number of paritions. It is clear that DWPA takes much lesser time due to much smaller number of variables in the local optimization problem (Feature dimensions for DSVM, number of paritions for DWPA). There is slight increase in time per iteration with increase in number of paritions due to increase in number of variables.

%
%

%


\section{Conclusion}
We propose a novel approach for training SVM in a distributed manner by learning an optimal set of weights for combining the SVM parameters independently
learnt on partitions of the entire dataset. Experimental results show that our method is much more accurate than \textit{parameter averaging} and 
is much faster than training SVM in feature space. Moreover, our method reaches an accuracy close to that of SVM trained in feature space in a much shorter 
time. We propose a novel proof to show that the stability final SVM parameter learnt using DWPA is $\mathcal{O}(\frac{1}{ML})$. Also, our method requires much
less network band-width as compared to DSVM when the number of features for a given dataset is very large as compared to the number of partitions, which is the 
usual scenerio for Big Data.

%
%
\bibliography{nips2014}
\bibliographystyle{plain}

 \section*{Appendix A}
 
 \begin{thm}
\label{theorem1app}
For any two arbitrary training samples of size $L$ differing by one sample point, the stability bound that holds for the parameter vectors returned
by support vector machine is:
\begin{equation}
\|\Delta \mathbf{w}\| \mbox{ is }\mathcal{O}(\frac{1}{ML}).
\end{equation}
\end{thm}

\textit{Proof.} 
Suppose we have two training datasets $S = (z_1,\cdots,z_{L-1},z_L)$ and $S' = (z_1,\cdots,z_{L-1},z'_L)$, where 
$\mathbf{z} = (\mathbf{x},y) \in \mathcal{X}\times\mathcal{Y}$, such that $\mathcal{X} \subset \mathbf{R}^d$ and   $\mathcal{Y} = \{-1,+1\}$. 
The two sets differ at a single data point: $\mathbf{z}_L = (\mathbf{x}_i,y_i)$ and $\mathbf{z}'_L = (\mathbf{x}'_i,y'_i)$.
Let $B_F$ be the Bregman divergence associated with a convex and non-diffentiable function $F$ defined for all $x,y$ by;
$B_F(x,y) = F(x) - F(y) - <g_y,(x - y)>$, where $g \in \partial F_y$ and $\partial F_y$ is the set of subdifferentials of $F$ at $y$. Since,
the minima is achieved at a point $y$ if $0 \in \partial F_y$. We define $g$ as follows;
\begin{equation}
\label{define_subgradient}
g =
\left\{
	\begin{array}{ll}
		0  & \mbox{if } 0 \in \partial F_y \\
		h  & \mbox{subject to,}\quad h \in  \partial F_y
	\end{array}
\right.
\end{equation}

Let $L_s:\mathbf{x} \rightarrow \sum_{i=1}^L H_{z_{i}}(\mathbf{x})$, where,
 $H_{z_{i}}(\mathbf{x}) = max(0,1-y\mathbf{w}^T\mathbf{x})$ denote the loss
  function and $G:\mathbf{x} \rightarrow \lambda\|\mathbf{x}\|^2$ denote the regularizer corresponding to the SVM problem. 
  Clearly, the function, $F_S = G + L_S$, is the objective function for SVM. $L_S$ is convex and non-differentiable while $G$ is convex and differentiable.
Since, Bregman divergence is non-negative ($B_{F} \geq 0 $), 
\begin{align}
& B_{F_{S}} = B_G +B_{L_{S}} \geq B_{G} \\
& B_{F_{S'}} \geq B_{G'}.
\end{align}

Thus, 
\begin{equation}
B_{G}(\mathbf{w'}\|\mathbf{w}) + B_{G}(\mathbf{w}\|\mathbf{w'}) \leq B_{F_{S}}(\mathbf{w'}\|\mathbf{w}) + B_{F_{S'}}(\mathbf{w}\|\mathbf{w'}).
\end{equation}

If $\mathbf{w}$ and $\mathbf{w'}$ are minimizers for of $B_{F_{S}}$ and
$B_{F_{S'}}$, then, $g_S(\mathbf{w}) = g_{S'}(\mathbf{w'}) = 0$ and
\begin{align}
\label{derivation1}
B_{F_{S}}(\mathbf{w'}\|\mathbf{w}) + B_{F_{S'}}(\mathbf{w}\|\mathbf{w'}) & = F_{S}(\mathbf{w'}) - F_{S}(\mathbf{w}) + F_{S'}(\mathbf{w}) - F_{S'}(\mathbf{w'}) \nonumber \\
& = \frac{1}{L}\left[ H_{{z}_L}(\mathbf{w'}) - H_{{z}_L}(\mathbf{w})  + H_{{z'}_L}(\mathbf{w}) - H_{{z'}_L}(\mathbf{w'})\right] \nonumber \\
& \leq -\frac{1}{L}\left[  g_{{z}_L}\cdot(\mathbf{w'})(\mathbf{w} - \mathbf{w'}) + g_{{z'}_L}\cdot(\mathbf{w})(\mathbf{w'} - \mathbf{w})  \right] \nonumber \\
& = -\frac{1}{L} \left[ g_{{z'}_L}(\mathbf{w}) - g_{{z}_L}(\mathbf{w'})\right]\cdot(\mathbf{w'} - \mathbf{w}) \nonumber \\
& = -\frac{1}{L} \left[ g_{{z'}_L}(\mathbf{w}) - g_{{z}_L}(\mathbf{w'})\right]\cdot(\Delta\mathbf{w})
\end{align}

From definition, 
\begin{equation}
\label{sumOfRegularizer}
B_{G}(\mathbf(w')\|\mathbf{w}) + B_{G}(\mathbf(w)\|\mathbf{w'}) = \lambda\|\Delta\mathbf{w}\|^2.
\end{equation}

Hence, from derivation ~\ref{derivation1} and equation ~\ref{sumOfRegularizer} and Cauchy-Schwarz inequality we have,
\begin{equation}
\lambda\|\mathbf{w}\| \leq \frac{1}{L}\| g_{{z'}_L}(\mathbf{w}) - g_{{z}_L}(\mathbf{w'}) \| \leq \frac{1}{L}\left[ \| g_{{z'}_L}(\mathbf{w})\| + \|g_{{z}_L}(\mathbf{w'}) \| \right]
\end{equation}

By definition, $H_{z_i}(\mathbf{w}) = max(0,1 - y_i\mathbf{w}^T\mathbf{x_i})$ and $g_{z_i}(\mathbf{w}) \in \partial H_{z_i}(\mathbf{w})$. Therefore,
\begin{align*}
& \partial H_{z_i}(w) \leq \| y_i \mathbf{x_i} \| \\
\Rightarrow & \partial H_{z_i}(w) \leq \|\mathbf{x_i} \| \\
\Rightarrow& \|g_{z_i}(\mathbf{w})\| \leq \|\mathbf{x_i}\|.
\end{align*}
 
If we assume that the feature vectors are bounded i.e., there exists a positive integer $R > 0$  such that for all training instances $(\mathbf{x},y) \in \mathcal{X} \times \mathcal{Y}$, $\|\mathbf{x} \| \leq R$, then we may state that,
\begin{equation}
\|\mathbf{w}\| \leq \frac{R}{\lambda L}
\end{equation}

Since, $\mathbf{w}$ is normalized and scaled by $\frac{1}{M}$. So, the bound on $\|\Delta \mathbf{w}\|$ in our case, is $\mathcal{O}(\frac{1}{ML})$.

\section*{Appendix B}

\begin{thm}
 \label{theorem:boundapp}
$\|\bm{\theta'} - \bm{\tilde{\theta}}\|$ is of the order of $\mathcal{O}(\frac{1}{ML})$.
\end{thm}

\textbf{Proof: } From definitions we have;
\begin{align}
\label{appC1}
 \|\bm{\theta'} - \bm{\tilde{\theta}}\|^2 &= \|\hat{W'}\bm{\beta'} - \hat{W'}\bm{\tilde{\beta}}\|^2\nonumber\\
 &\leq \|\bm{\beta'} - \tilde{\bm{\beta}}\|_K^2, \mbox{ where ,}K = \hat{W'}^T\hat{W'}
\end{align}

Since, we have a lower bound on $\|\bm{\beta'} - \tilde{\bm{\beta}}\|^2 \leq\frac{1}{\bm{\sigma}_{min}}\|\bm{\beta'} - \tilde{\bm{\beta}}\|_K^2$,
where $\sigma_{min}$ is the minimum eigenvalue of $K$.

Hence, we need to prove an upper bound on $\|\bm{\beta'} - \tilde{\bm{\beta}}\|$.

From the reasoning of theorem \ref{theorem1}, we have;
\begin{equation}
 B_G(\bm{\beta'}\|\tilde{\bm{\beta}}) + B_G(\tilde{\bm{\beta}}\|\bm{\beta'}) \leq B_{F_S}(\bm{\beta'} \| \bm{\tilde{\beta}}) + B_{F_S'}(\bm{\tilde{\beta}} \| \bm{\beta'})
\end{equation}

From the \textit{left hand side} of the equation we have;
\begin{equation}
 B_G(\bm{\beta'}\|\tilde{\bm{\beta}}) + B_G(\tilde{\bm{\beta}}\|\bm{\beta'}) = (\tilde{\bm{\beta}} + \bm{\beta'})^T(\hat{W'}^T\hat{W'})( \bm{\beta'} - \tilde{\bm{\beta}})
\end{equation}

From the \textit{right hand side} of the equation, using the similar reasoning as that used for $\|\bm{\beta} - \tilde{\bm{\beta}}\|$, we have;
\begin{align}
 B_{F_S}(\bm{\beta'} \| \bm{\tilde{\beta}}) + B_{F_S'}(\bm{\tilde{\beta}} \| \bm{\beta'}) &= L_S(\bm{\beta'}) - L_S(\bm{\tilde{\beta}}) + L_{S'}(\bm{\tilde{\beta}}) - L_{S'}({\bm{\beta}})\nonumber\\
 &\leq \frac{1}{ML}[max(0,(\hat{W'}\bm{\beta'})^T(y_{z'_m}\bm{x}_{z'_m} - y_{z_m}\bm{x}_{z_m}) - (\hat{W'}\bm{\tilde{\beta}})^T(y_{z'_m}\bm{x}_{z'_m} - y_{z_m}\bm{x}_{z_m})]\nonumber\\
 &\leq \frac{1}{ML}|\hat{W'}(\bm{\tilde{\beta}} - \bm{\beta'})(y_{z'_m}\bm{x}_{z'_m} - y_{z_m}\bm{x}_{z_m})|\nonumber\\
 &\leq \frac{2R}{ML}\|\bm{\tilde{\beta}} - \bm{\beta'}\|
\end{align}
Equating, \textit{left hand side} and \textit{right hand side} of the equation, we get;
 $\|\bm{\beta'} - \tilde{\bm{\beta}}\|$ is $\mathcal{O}(\frac{1}{ML})$, and hence, from \ref{appC1}, we get, $\|\bm{\theta'} - \bm{\tilde{\theta}}\|$ 
 is $\mathcal{O}(\frac{1}{ML})$.
\section*{Appendix C}
%
  \begin{figure}[ht] 
   \captionsetup[subfigure]{labelformat=empty}
  \begin{center}
   \subfloat[]{
   \includegraphics[width=0.34\linewidth]{./plot/primal/r_50}
    }
   \hspace{-.45cm}
    \subfloat[]{
   \includegraphics[width=0.34\linewidth]{./plot/primal/r_100}
    }
   \hspace{-.45cm}
    \subfloat[]{
   \includegraphics[width=0.34\linewidth]{./plot/primal/r_200}
    }
    \hspace{-.45cm}
   \subfloat[][50 partitions]{
   \includegraphics[width=0.34\linewidth]{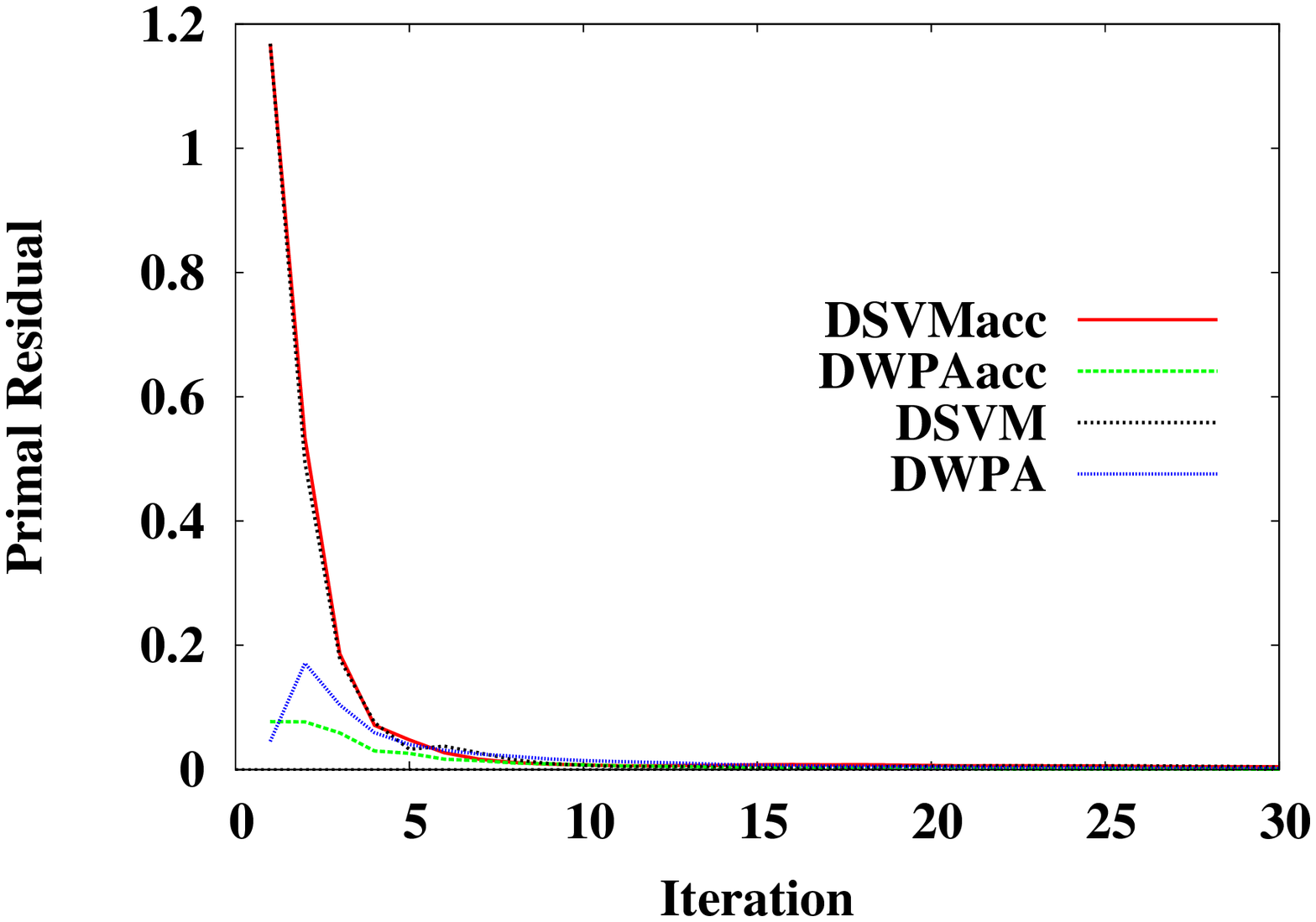}
    }
   \hspace{-.45cm}
    \subfloat[][100 partitions]{
   \includegraphics[width=0.34\linewidth]{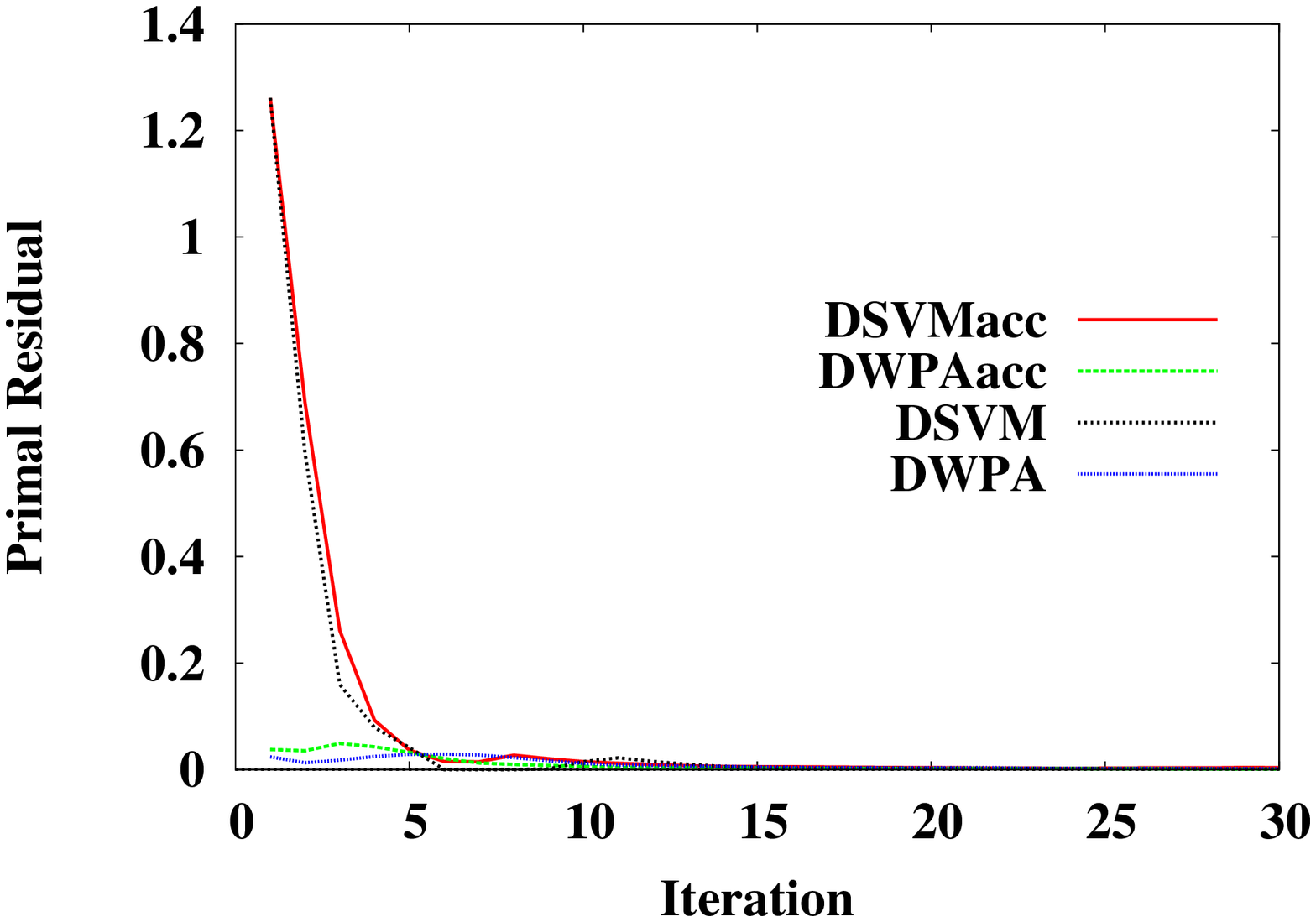}
    }
  \hspace{-.45cm}
    \subfloat[][200 partitions]{
   \includegraphics[width=0.34\linewidth]{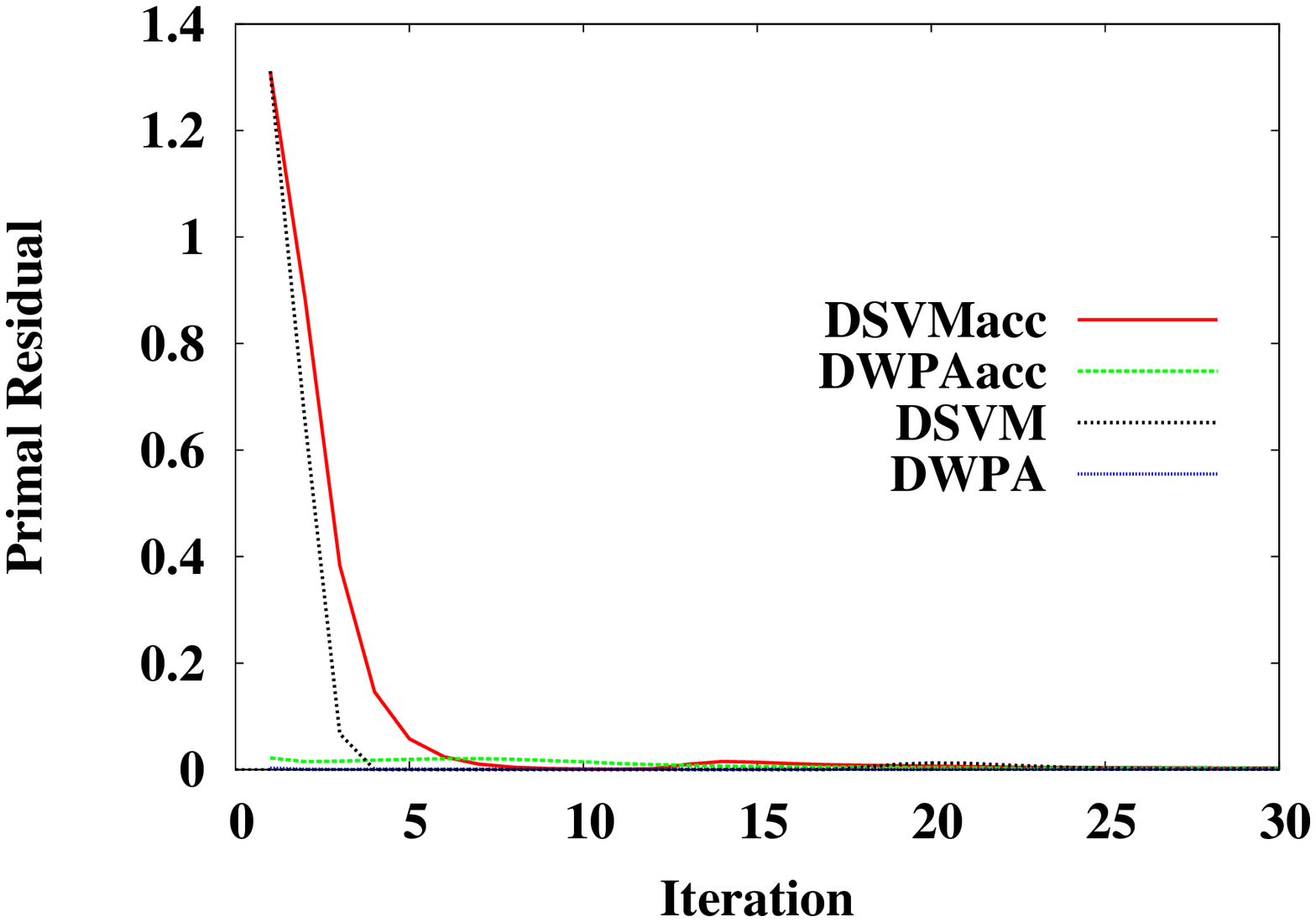}
   }
  \hspace{-.45cm}
   \subfloat[][50 partitions]{
   \includegraphics[width=0.34\linewidth]{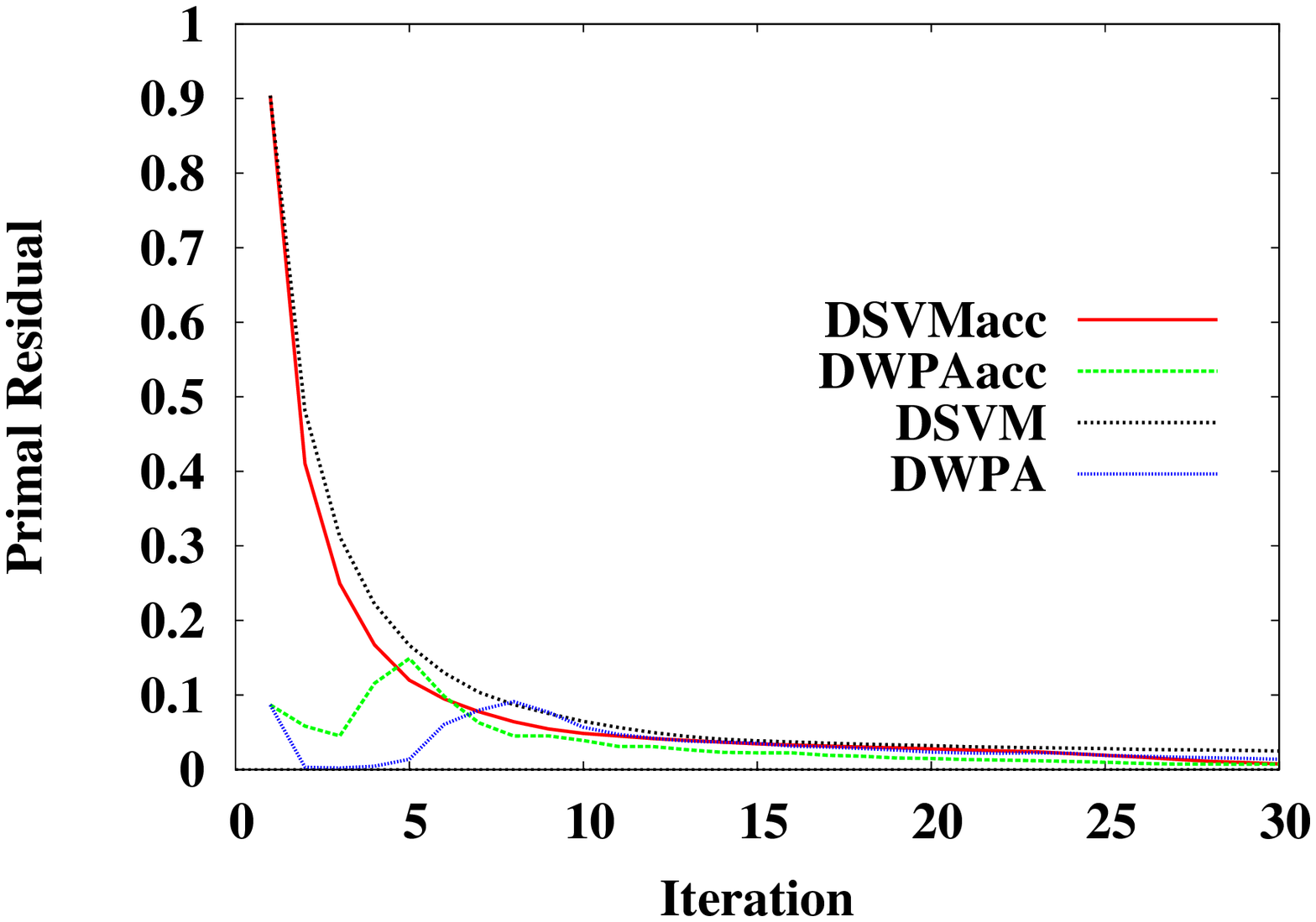}
    }
   \hspace{-.45cm}
    \subfloat[][100 partitions]{
   \includegraphics[width=0.34\linewidth]{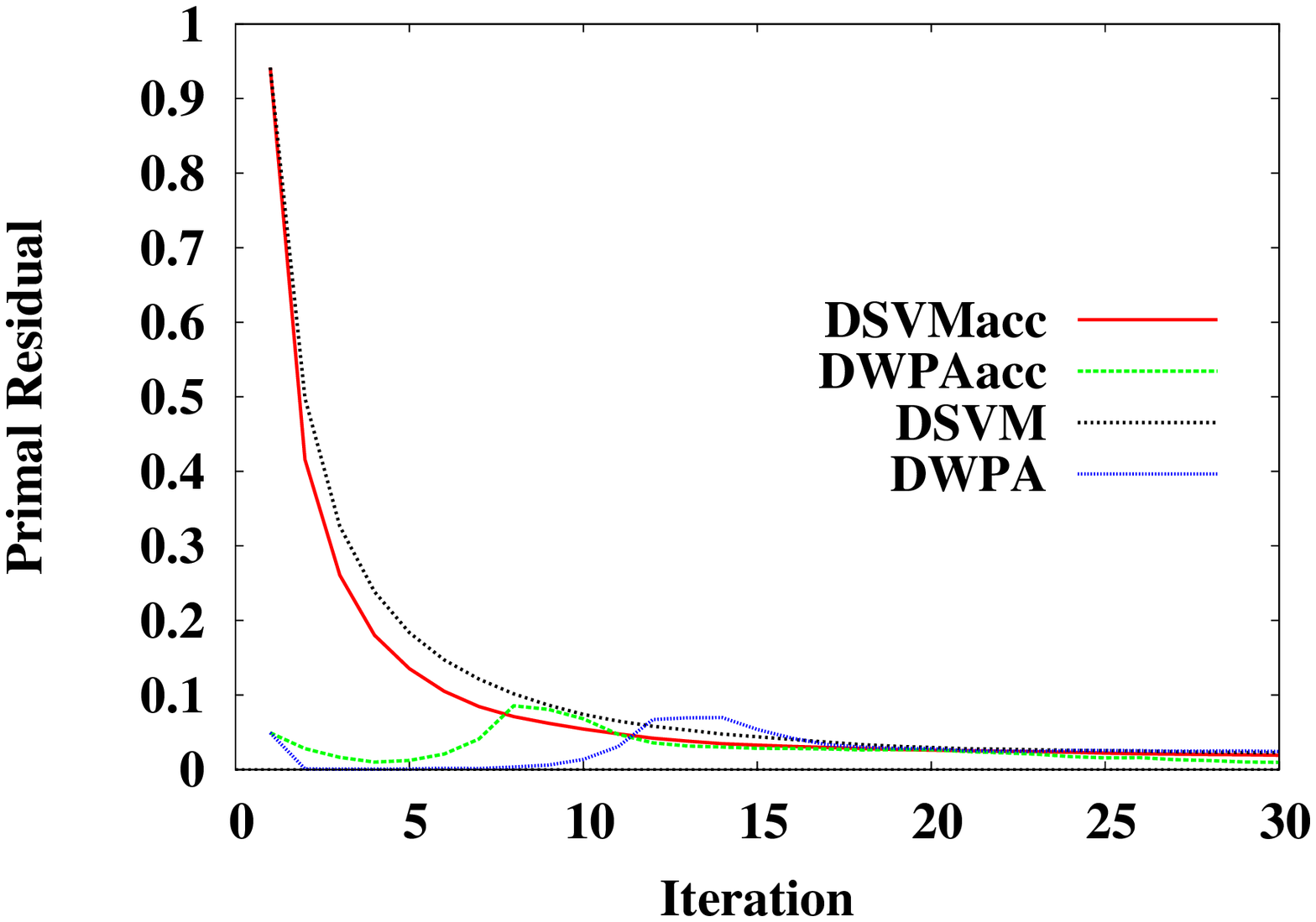}
    }
  \hspace{-.45cm}
    \subfloat[][200 partitions]{
   \includegraphics[width=0.34\linewidth]{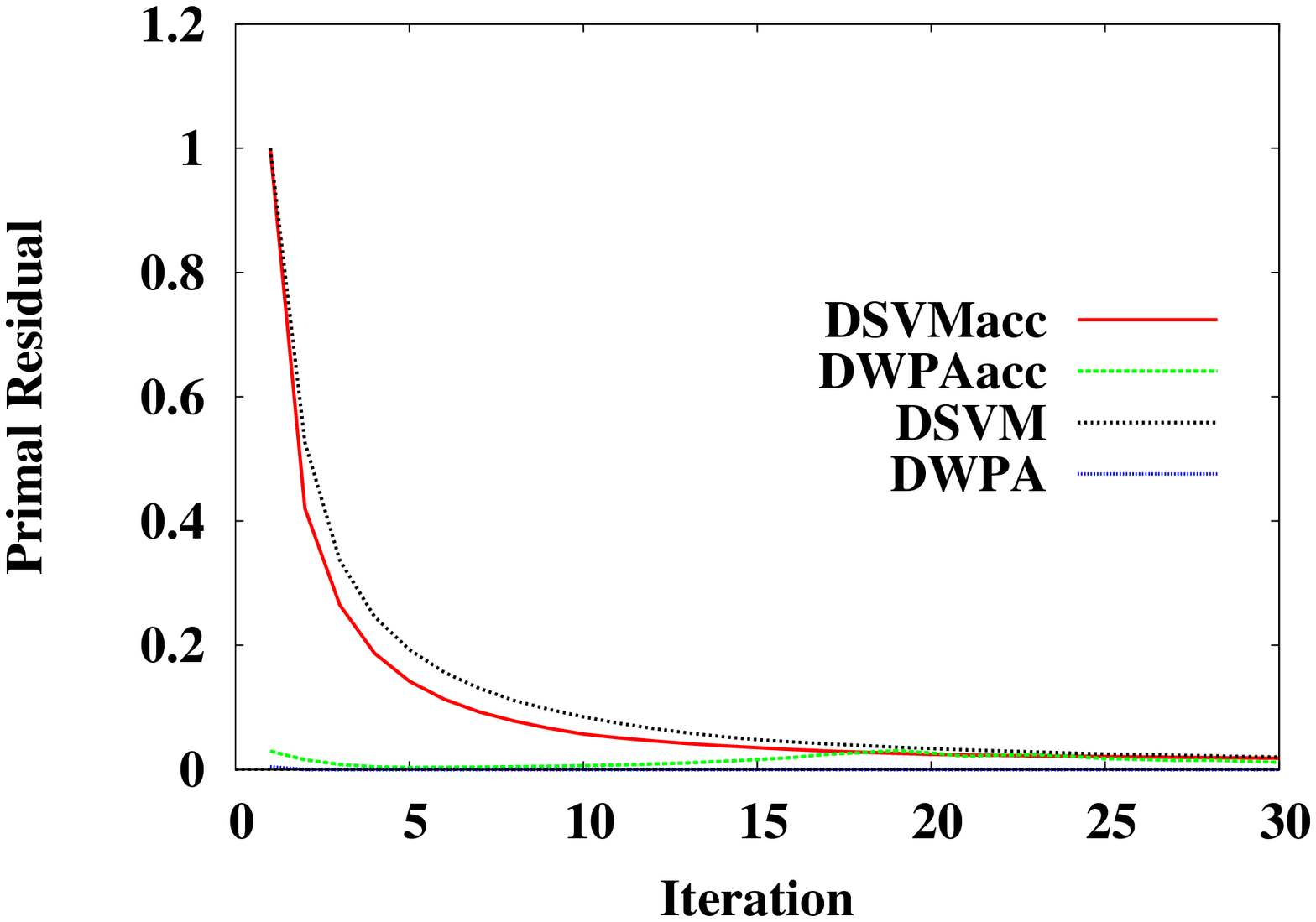}
   }   
 \end{center}
   \caption{\label{pe}Convergence of primal residual for \textbf{real-sim}(top), \textbf{epsilon}(middle) and \textbf{gisette}(bottom)}
 \end{figure}
%
%

   \begin{figure}[ht] 
   \captionsetup[subfigure]{labelformat=empty}
  \begin{center}
   \subfloat[]{
   \includegraphics[width=0.34\linewidth]{./plot/testacc/r_50}
    }
   \hspace{-.45cm}
    \subfloat[]{
   \includegraphics[width=0.34\linewidth]{./plot/testacc/r_100}
    }
    \hspace{-.45cm}
    \subfloat[]{
   \includegraphics[width=0.34\linewidth]{./plot/testacc/r_200}
    }
    \hspace{-.45cm}
   \subfloat[][50 partitions]{
   \includegraphics[width=0.34\linewidth]{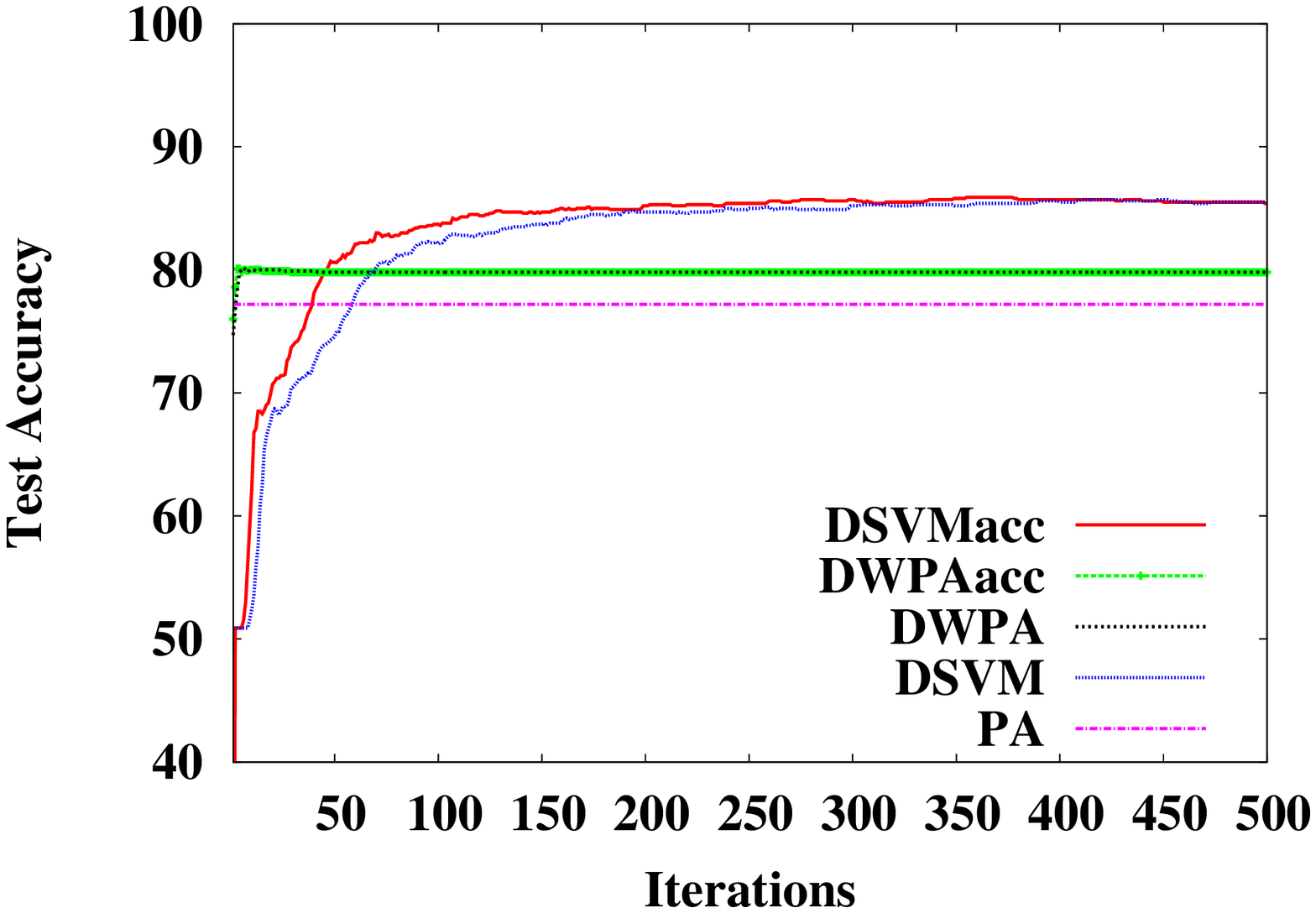}
    }
   \hspace{-.45cm}
    \subfloat[][100 partitions]{
   \includegraphics[width=0.34\linewidth]{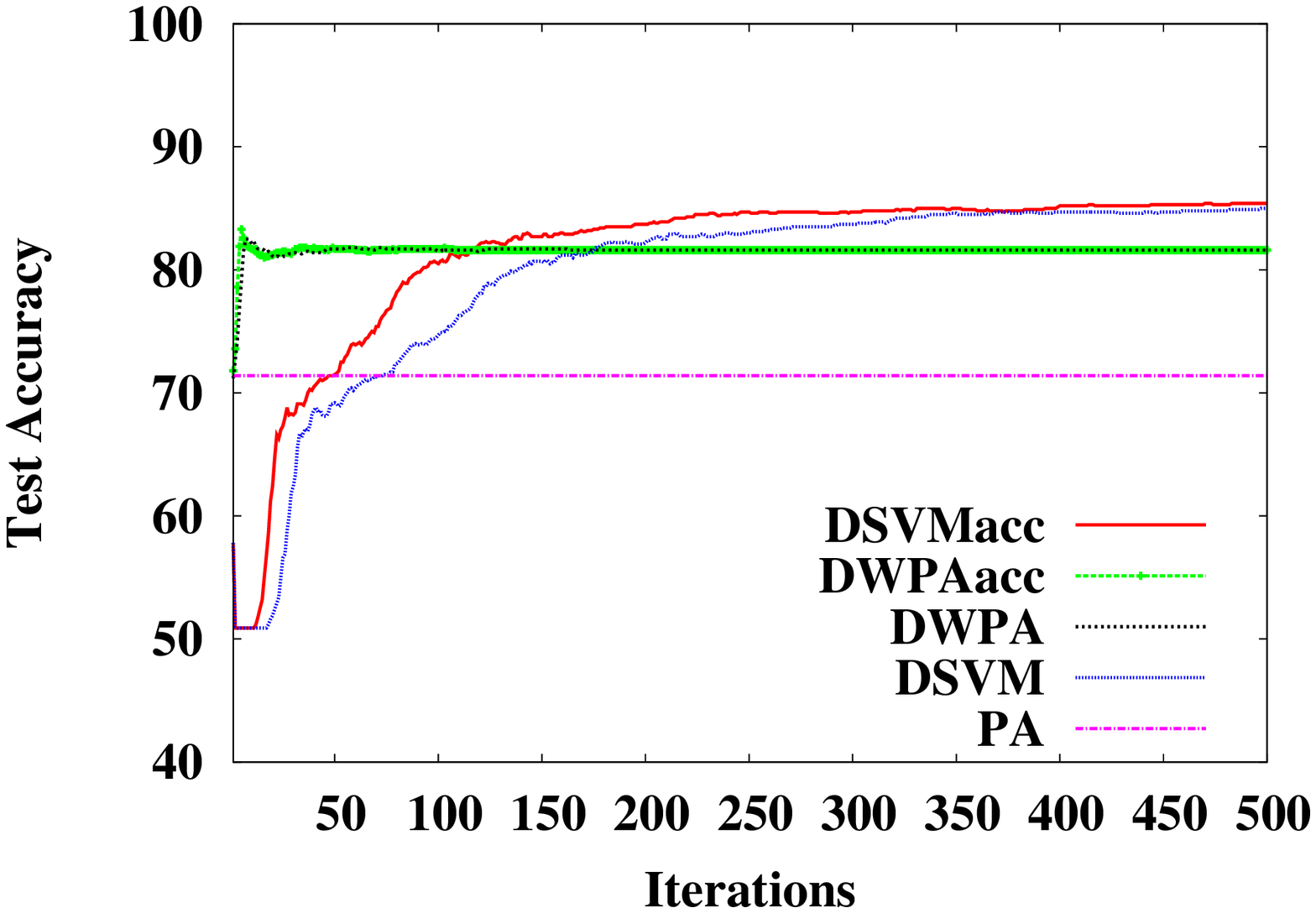}
    }
   \hspace{-.45cm}
    \subfloat[][200 partitions]{
   \includegraphics[width=0.34\linewidth]{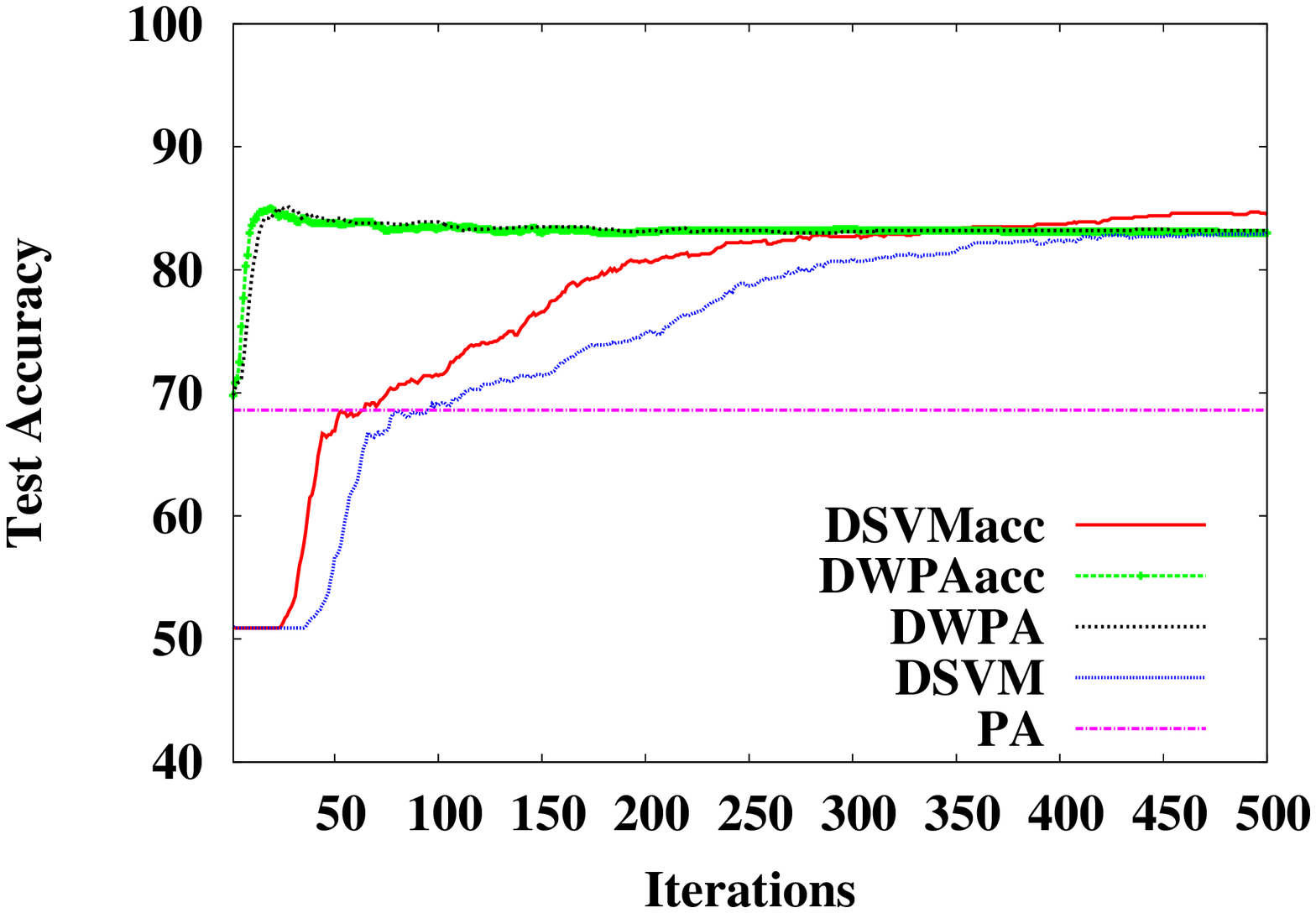}
    }
        \hspace{-.45cm}
   \subfloat[][50 partitions]{
   \includegraphics[width=0.34\linewidth]{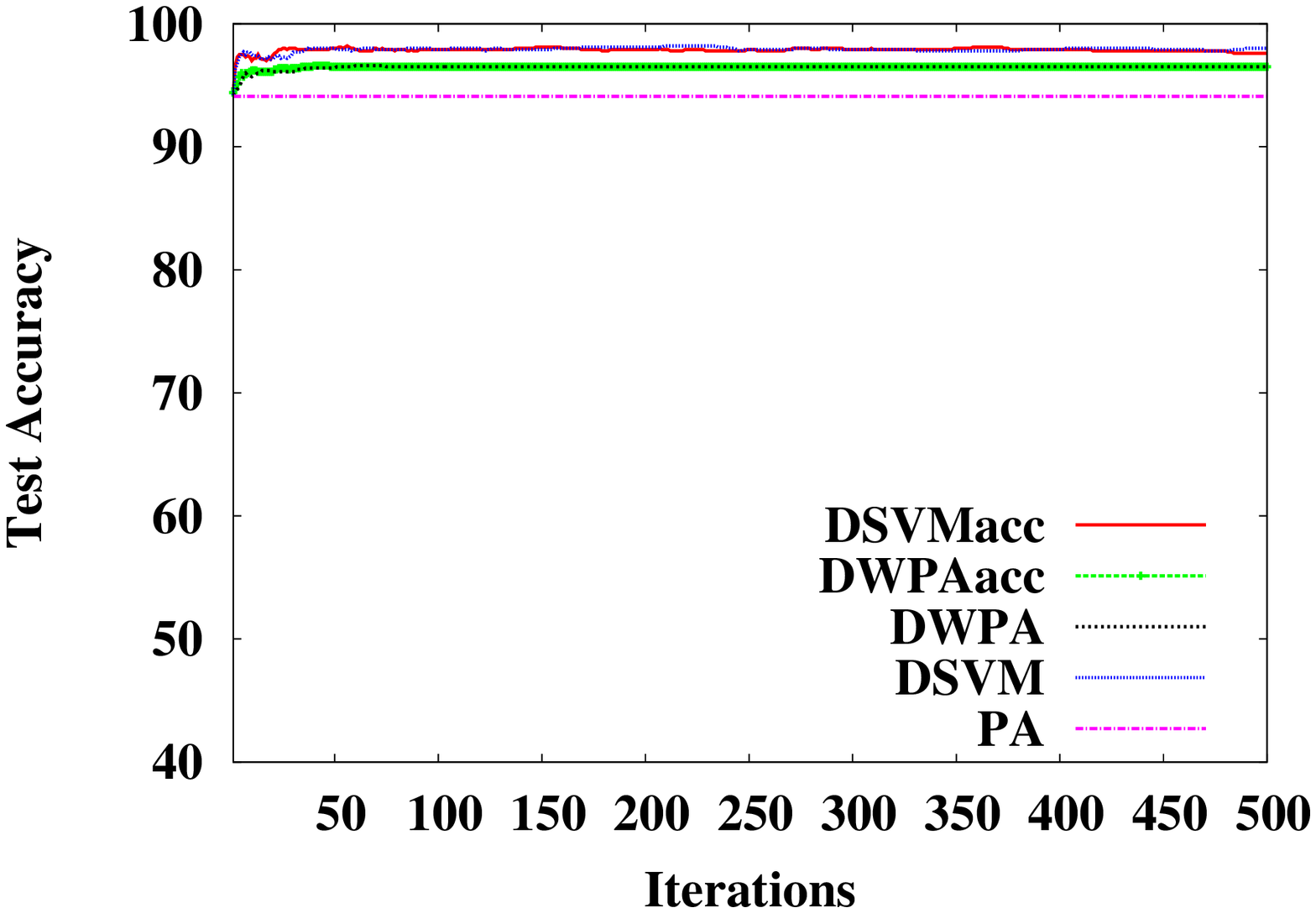}
    }
   \hspace{-.45cm}
    \subfloat[][100 partitions]{
   \includegraphics[width=0.34\linewidth]{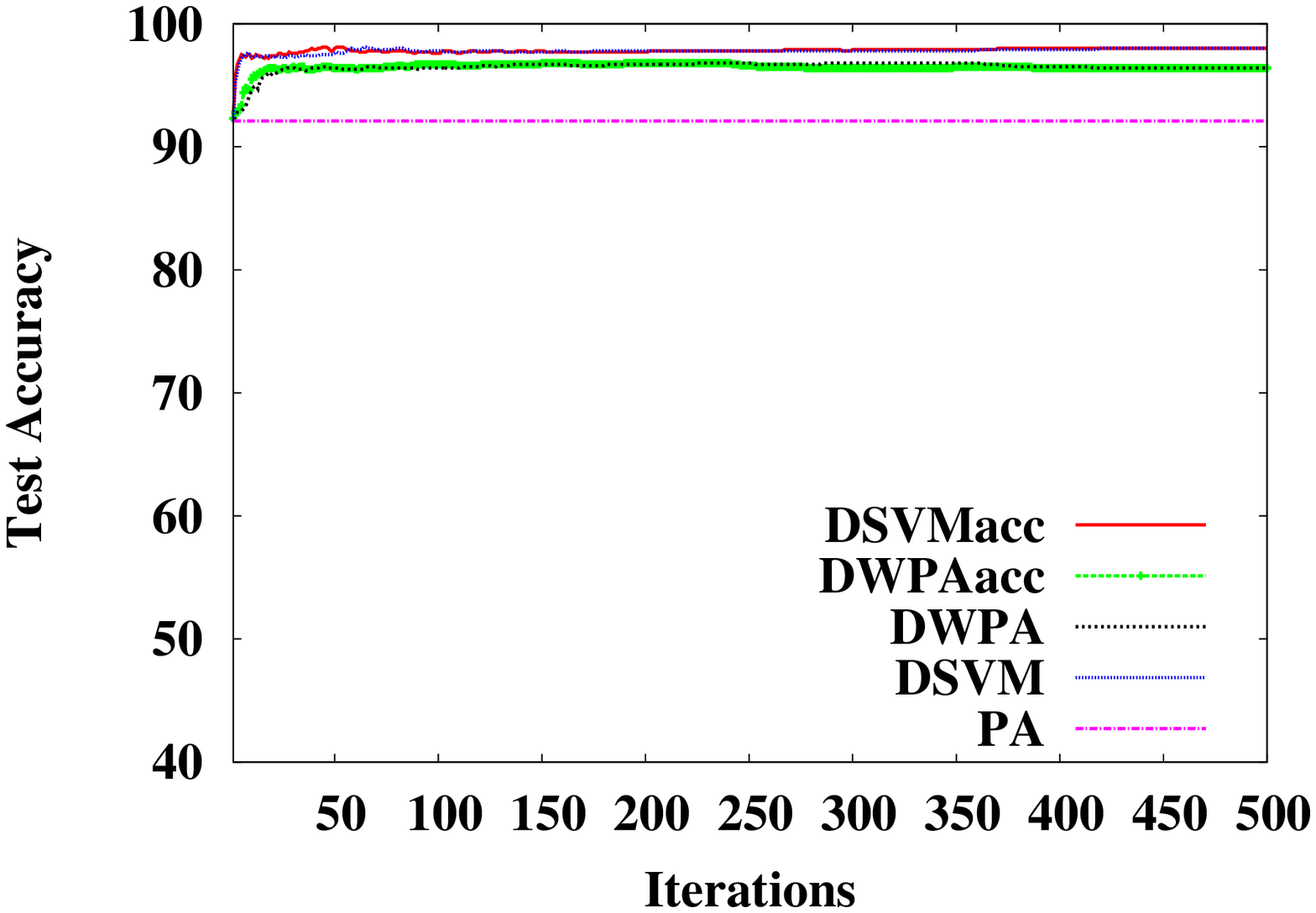}
    }
   \hspace{-.45cm}
    \subfloat[][200 partitions]{
   \includegraphics[width=0.34\linewidth]{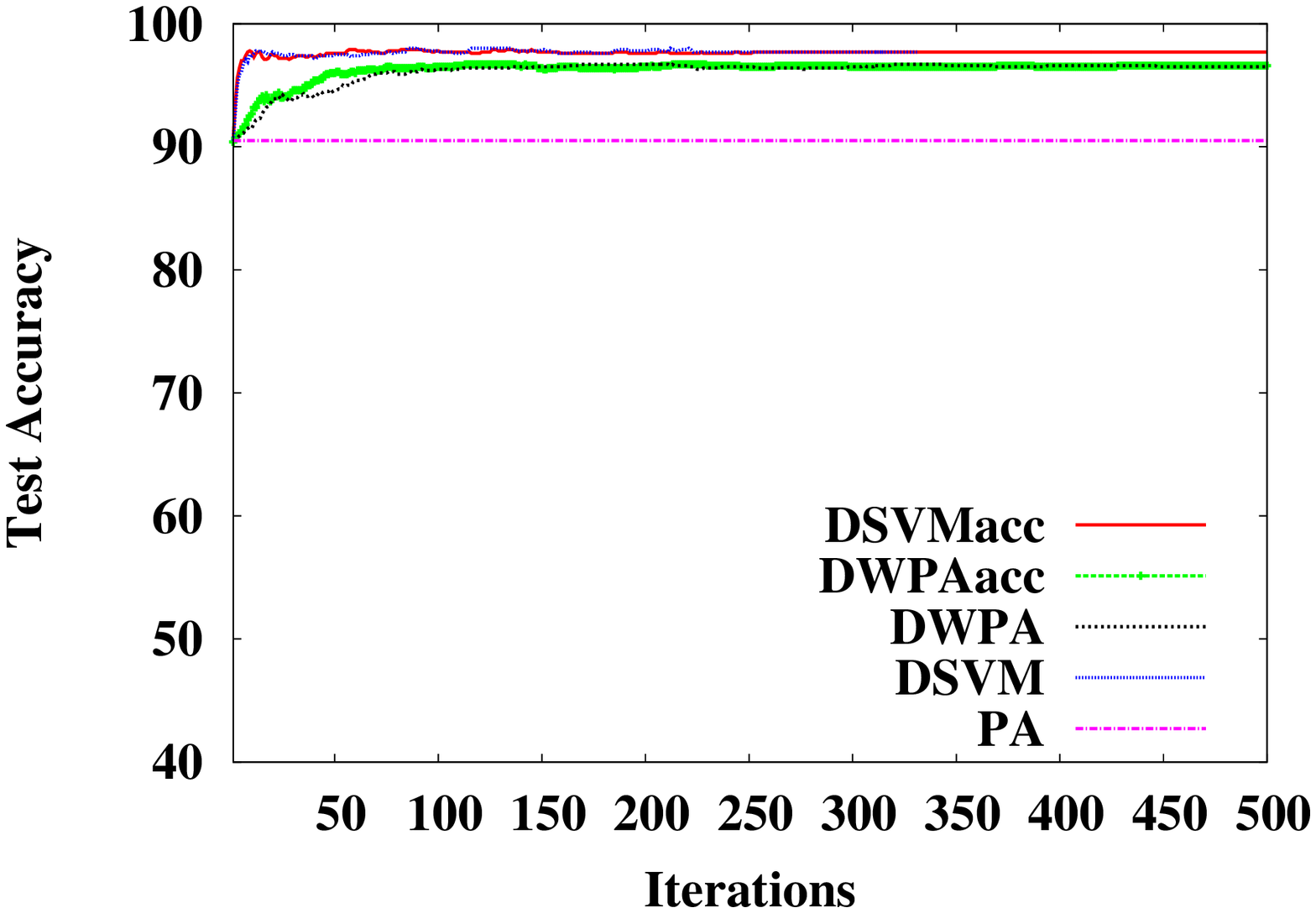}
    }
 \end{center}
   \caption{\label{te}Testset accuracies for \textbf{real-sim}(top), \textbf{epsilon}(middle) and \textbf{gisette}(below)}
 \end{figure}

\end{document}